\documentclass[twocolumn, switch]{article} 

\usepackage{preprint}

\usepackage{amsmath, amsthm, amssymb, amsfonts}
\usepackage{bm}
\usepackage{mathtools}

\usepackage[round]{natbib}
\usepackage[utf8]{inputenc}	
\usepackage[T1]{fontenc}	
\usepackage{xcolor}		
\usepackage{hyperref}
\hypersetup{colorlinks = true,
            linkcolor = blue,
            urlcolor  = blue,
            citecolor = blue,
            anchorcolor = black
}
\usepackage{booktabs} 		
\usepackage{nicefrac}		
\usepackage{microtype}		
\usepackage{lineno}		
\usepackage{float}			

\usepackage{caption}
\usepackage{array}
\usepackage[ruled,vlined]{algorithm2e}

\usepackage{newfloat}
\DeclareFloatingEnvironment[name={Supplementary Figure}]{suppfigure}

\usepackage{titlesec}
\titlespacing\section{0pt}{12pt plus 3pt minus 3pt}{1pt plus 1pt minus 1pt}
\titlespacing\subsection{0pt}{10pt plus 3pt minus 3pt}{1pt plus 1pt minus 1pt}
\titlespacing\subsubsection{0pt}{8pt plus 3pt minus 3pt}{1pt plus 1pt minus 1pt}

\usepackage{graphics}
\usepackage{multirow}
\usepackage{hhline}
\usepackage{subfigure}

\title{Learning Compact Reward for Image Captioning}

\usepackage{authblk}

\author{Nannan Li}
\author{Zhenzhong Chen\thanks{\tt{zzchen@ieee.org}}}
\affil{School of Remote Sensing and Information Engineering, Wuhan University}

\begin{document}

\twocolumn[ 
  \begin{@twocolumnfalse} 
  
\maketitle

\begin{abstract}
Adversarial learning has shown its advances in generating natural and diverse descriptions in image captioning. However, the learned reward of existing adversarial methods is vague and ill-defined due to the reward ambiguity problem. In this paper, we propose a refined Adversarial Inverse Reinforcement Learning (rAIRL) method to handle the reward ambiguity problem by disentangling reward for each word in a sentence, as well as achieve stable adversarial training by refining the loss function to shift the generator towards Nash equilibrium. In addition, we introduce a conditional term in the loss function to mitigate mode collapse and to increase the diversity of the generated descriptions. Our experiments on MS COCO and Flickr30K show that our method can learn compact reward for image captioning.
\end{abstract}
\vspace{0.35cm}

  \end{@twocolumnfalse} 
] 



\section{Introduction}
\label{intro}
\begin{figure}[t]
\centering
   \includegraphics[width=0.45\textwidth]{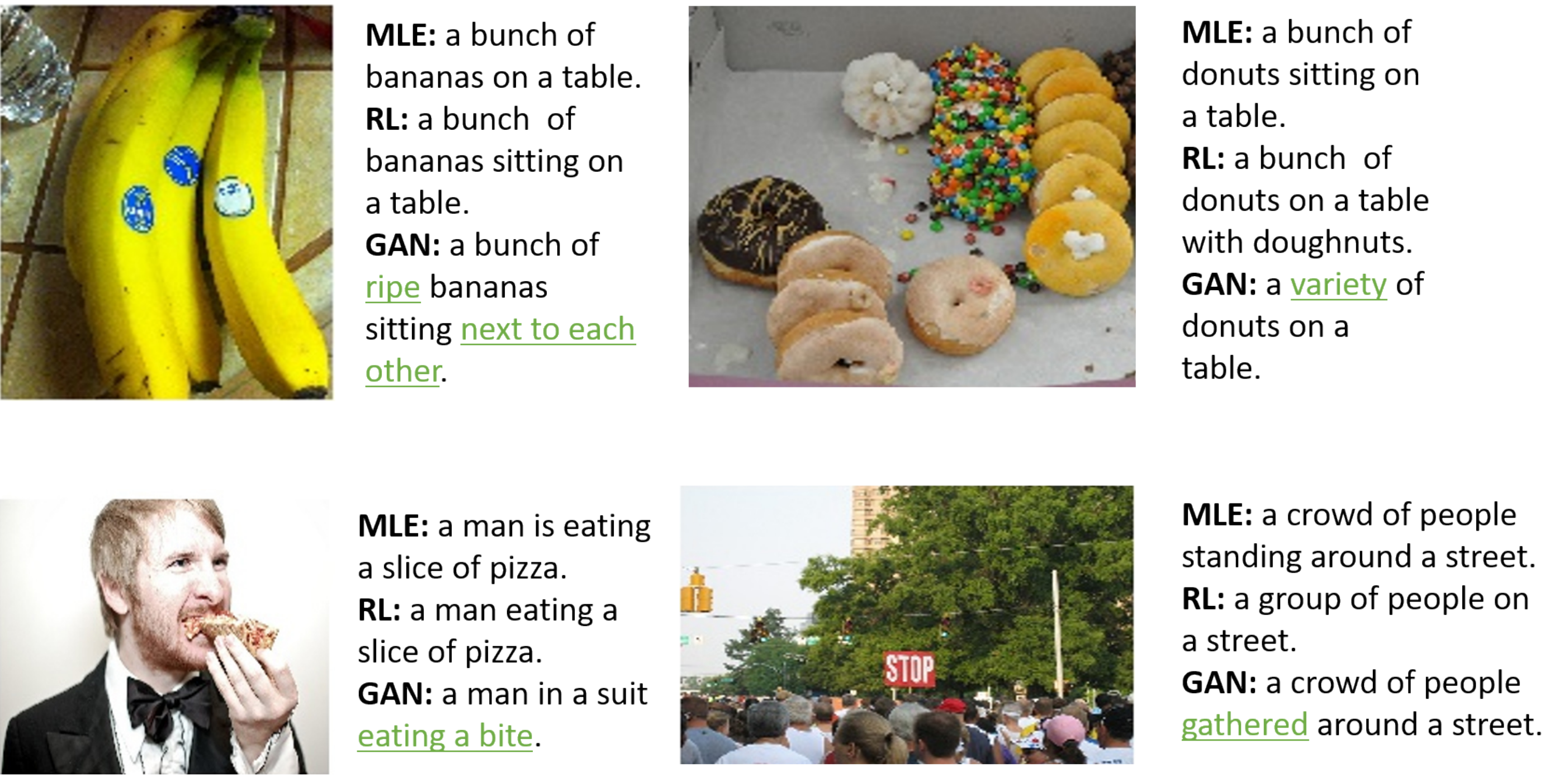} 
  \caption{\small{Examples of the generated captions on MS COCO. Compared with MLE and RL, GAN produces captions with diverse forms and human-like descriptions.}}
  \label{fig:exp_diff} 
\end{figure}

Image captioning is a task of generating descriptions of a given image in natural language. In a general encoder-decoder structure \citep{Vinyals2015Show}, image features are encoded in a CNN and decoded into a caption in a word by word manner. Based on the loss function, typical approaches addressing the problem could be divided into three categories: MLE (Maximum Likelihood Estimation), RL (Reinforcement Learning) and GAN (Generative Adversarial Network).\\
\indent
Early proposed methods were based on MLE function and made improvements by designing specific model structure \citep{xu2015show}. MLE adopts the cross-entropy loss and learns a one-hot distribution for each word in the sentence. By maximizing the probability of the ground truth word whilst suppressing other reasonable vocabularies, the probability distribution learned by MLE tends to be \emph{sparse} and the generated captions have limited diversity \citep{DaiLUF17}. On the other hand, RL has advantages in boosting the model performance by optimizing the handcrafted metrics \citep{Rennie2016Self,si2017improve,chen2019improving}. However, due to the reward hacking problem, RL maximizes the reward in an unintended way and fails to produce human-like descriptions \citep{me}. Considering naturalness and diversity of the generated captions, GAN has raised attention in image captioning for its capability of producing descriptions that are indistinguishable from human-written ones \citep{DaiLUF17,speak,chen2019improving,DogninImproved}. See Figure \ref{fig:exp_diff} for a few examples.\\
\begin{figure}[t]
\centering
    \includegraphics[width=0.45\textwidth]{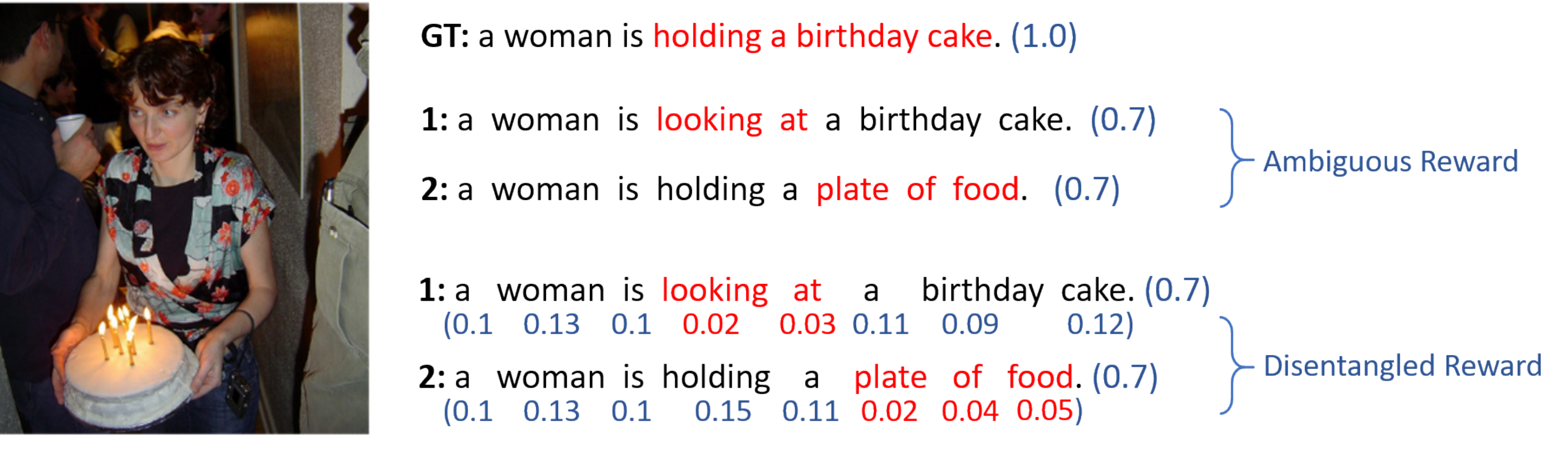} 
  \caption{\small{An example showing the difference between ambiguous reward and disentangled reward. GT represents the ground truth caption. Number beside each caption is its learned reward. Ambiguous reward function may give the same reward for two different captions, but which word(s) causes the reward to increase or decrease is not accounted for. Disentangled reward function provides word-wise reward and can further locate the wrong words.}}
\label{fig:exp}
\end{figure}
\indent
In image captioning, the generator of GAN learns true data distribution by maximizing the reward function learned from a discriminator, and the discriminator distinguishes the generated sample from the true data. The adversarial training converges to an equilibrium point (i.e., Nash equilibrium) at which both the generator and discriminator cannot improve \citep{goodfellow2014generative}. GAN is less biased towards frequently occurring n-grams and learns to describe images with human-like descriptions \citep{speak}. However, previous work of adversarial networks in image captioning gives one reward function $D$ for a complete sentence consisting of $n$ words. This strategy causes the reward ambiguity problem \citep{ng1999} since which word(s) causes the reward to increase or decrease is not accounted for, and thus there are many optimal policies that determine the sentence can explain one reward. As shown in Figure \ref{fig:exp}, the generated two captions have the same reward ($0.7$) in GAN, whereas the contribution of each word to this reward remains unknown. The first caption gives the wrong verb ``looking at'' whilst the second caption has an incorrect object ``a plate of food''. However, the ambiguous reward in GAN makes it unable to locate the inappropriate words. On the other hand, from the perspective on the system level, learning sentence-level reward from different image-caption pairs is analogous to learning reward of a trajectory from different system dynamics, which makes the discriminator unable to distinguish the true reward functions from those shaped by the environment dynamics \citep{Fu2018LearningRobust}.\\
\indent
Facing above challenges, we adopt Adversarial Inverse Reinforcement Learning (AIRL) \citep{Fu2018LearningRobust} to solve the reward ambiguity problem by learning a \emph{compact} reward function, where \emph{compact} means the reward function should satisfy two conditions: 1) The reward is word-wise and disentangled for each word in a sentence from different image-caption pairs, as shown in Figure \ref{fig:exp}.  2) The reward difference of two words is positively correlated to their semantic difference. For instance, words with similar semantics, such as \emph{children} and \emph{kids}, correspond to close reward values. A compact reward function can precisely tell the contribution of each word and thus help to locate the wrong words. It saves the effort of predefining a handcrafted reward function, and can recover the true reward up to a constant at optimality. Driven by such compact reward function from the discriminator, the generator can learn the optimal policy and thus produces qualitative descriptions. However, there are still two major problems to address: 1) AIRL is difficult to converge to Nash equilibrium using policy gradient, requiring Hessian of the gradient vector filed being positive definite \citep{the-numerics-of-gans}. We will discuss this in detail in Section \ref{sec:gen}. 2) As a GAN based method, AIRL has a sharp decision boundary for two disjoint distributions, which means the discriminator can be far more stronger than the generator. The consequence is a limited diversity in the generated captions, which is a commonly encountered issue in GAN called mode collapse \citep{MirzaO14}.\\
\indent
In this paper, we propose a refined AIRL method to learn a compact reward function for each word, as well as achieve stable adversarial training by refining the loss function to shift the generator towards Nash equilibrium. In addition, a conditional term is introduced in the loss function to mitigate mode collapse and to increase the diversity of the generated descriptions. Both the caption evaluator (i.e., discriminator) \citep{cui2018learning, sharif2018learning-based} and the generator are cast into this unified framework, where the discriminator evaluates captions using a learned compact reward function, and the generator produces qualitative image descriptions. We demonstrate the effectiveness of our method in the experiments. 
%

\section{Related Work}
\paragraph{Image captioning.} The development of image captioning can be summarized into two directions: model structure design \citep{Lu2017Knowing, explore2018yao} and loss function construction \citep{Rennie2016Self, Ren_2017_CVPR}. In the methods based on model structure design, attention mechanism and the fusion of visual and semantic information are the key focus. \citet{Lu2017Knowing, Lu2018Neural} proposed a sentinel gate to learn adaptive attention between visual content and non-visual text. \citet{explore2018yao} explored the role of visual relationship in image captioning. On the other hand, methods based on loss function construction focus on optimization of the loss function. \citet{Rennie2016Self} optimized on non-differentiable evaluation metric using policy gradient, and improved scores of these metrics on various models. \citet{Ren_2017_CVPR} designed an embedding reward under actor-critic reinforcement learning. Similarly, we address the construction of loss functions, and thus our algorithm can be built on existing model structures.\\
\indent
\paragraph{Adversarial Methods for Image Captioning. }  Adversarial methods are known for producing plausible samples by training the generator and the discriminator in an adversarial manner \citep{goodfellow2014generative}. In image captioning, the discriminator is formed as a binary classifier that distinguishes the generated sentence from the ground truth, while the generator produces captions that can fool the discriminator. Conditional GAN was proposed in \citep{DaiLUF17} to improve the naturalness and diversity of generated captions. CNN and RNN based discriminators were introduced in \citep{chen2019improving}. However, existing methods estimate a reward function for the complete sentence consisting of $n$ words, where multiple optimal policies that determine the sentences can correspond to one reward \citep{ng1999}. Thus the learned reward is ambiguous and ill-defined. We solve this problem by recovering a compact reward function for each word in the sentence under a refined AIRL framework. Although AIRL has been utilized to solve problems in other fields \citep{WangNo18, ziming19diag, zhan18towards}, we are the first to make algorithmic improvements to AIRL such that the model can converge to Nash equilibrium, and that diversity of the outputs can be increased.

\section{Adversarial Inverse Reinforcement Learning}
\label{sec:adrl}
Due to the high variance estimate of a full sentence and the reward ambiguity problem, instead of learning reward for a complete sentence, we could learn the reward distribution ${p_\theta}(w_t,s_t)$ at time $t$ for each word-state pair $(w_t,s_t)$ so that the true reward function can be recovered at optimality \citep{Fu2018LearningRobust}. In the following, we use $w_t$ to represent the word at time $t$, and $s_t$ is the corresponding state vector at time $t$. Note that in an LSTM based model structure, $s_t$ refers to the hidden state of the LSTM cell. In the following, we introduce how AIRL disentangles reward for each word-state pair $(w_t,s_t)$.\\
\indent
AIRL is an adversarial reward learning algorithm based on Maximum-Entropy-IRL. \citet{Finn2016AConnection} first proved that  Maximum-Entropy-IRL is mathematically equivalent to GAN under a special form of the discriminator:
\begin{equation}
{D_{\theta} }(w_t,s_t) = \frac{{{p_\theta }(w_t,s_t)}}{{{p_\theta }(w_t,s_t) + {{\pi}_{\psi}} (t)}}
\end{equation}
where ${p_\theta}(w_t,s_t)$ is the data distribution estimated by the discriminator at time $t$, parameterized by $\theta$. ${p_\theta}(w_t,s_t)$ is estimated using the natural exponential function ${p_\theta}(w_t,s_t)= \text{exp}\left\{ {{f_\theta }(w_t,s_t)} \right\}$, where ${{f_\theta }(w_t,s_t)}$ is the reward function. ${{\pi}_{\psi}} (t)$ is the policy distribution produced by the generator at time $t$, paramterized by $\psi$. ${{\pi}_{\psi}} (t)$ is the generated vocabulary distribution under the context of image captioning. $D$ is the decision boundary, which represents the probability that $(w_t,s_t)$ comes from the true word distribution rather than ${{\pi}_{\psi}}$. The discriminator is trained to differentiate between the true words and the generated words whereas the generator tries to fool the discriminator by learning a policy ${{\pi}_{\psi}}$ to maximize the reward $f_\theta$ from $D$.\\
\indent
Considering the reward ambiguity problem, \citet{Fu2018LearningRobust} further extended the above theory to AIRL by introducing a reward shaping term $h_{\varphi}$ into $f_{\theta}(w_t,s_t)$. The reward shaping term disentangles reward from different system dynamics, which refer to different image-caption pairs under the context of image captioning.
\begin{equation}
{f_{\theta ,\varphi }}(w_t,s_t) = {g_\theta }(w_t,s_t;{s_{t+1}}) + \gamma {h_\varphi }({{s_{t+1}} }) - {h_\varphi }(s_t)
\label{eq:3}
\end{equation}
where ${g_\theta}$ denotes the reward approximator that recovers the true reward up to a constant, and $h_{\varphi}$ is the reward shaping term that preserves the optimal $\pi_\psi$. $\gamma$ is a constant in range $(0,1]$.\\
\begin{algorithm*}[!t]
\caption{refined AIRL}
\SetAlgoNoLine
\DontPrintSemicolon
Initialize the vocabulary distribution $\pi_{\psi}$ and discriminator ${f_{\theta ,\varphi }}$.\;
\For{$\mathrm{iteration}$ $i$ $\mathrm{in}$ $\{1,...,N\}$}{
Obtain caption $\{w_1^\text{true},...,w_n^\text{true}\}$ from the ground truth.\;
Collect generated caption $\{w_1,...,w_n\}$ using the vocabulary distribution ${\pi_{\psi}}(t)$.\;
$D_{\theta,\varphi} {\leftarrow} {\mathrm{sigmoid}}({f_{\theta ,\varphi }}- \log ({\pi_{\psi}}(t) ))$\;
Update $(\theta ,\varphi)$ via Eq. (\ref{eq:dis}) for the discriminator.\;
Update $\psi$ via Eq. (\ref{eq:gen}) for the generator.\;
  }
\end{algorithm*}
\indent
Then the estimated data distribution becomes
\begin{equation}
{p_{\theta ,\varphi }}(w_t,s_t)= \text{exp}\left\{ {{f_{\theta ,\varphi }}(w_t,s_t)} \right\}
\end{equation}
\indent
For convenience, the decision boundary $D$ can be represented as a sigmoid function:
\begin{equation}
\begin{aligned}
  {D_{\theta ,\varphi }}({w_t},{s_t}) &= \frac{{{p_{\theta ,\varphi }}({w_t},{s_t})}}{{{p_{\theta ,\varphi }}({w_t},{s_t}) + {\pi _\psi }(t)}} \\ 
   &= {\text{sigmoid}}\left ({f_{\theta ,\varphi }}\left ({w_t},{s_t}\right ) - \log \left ({\pi _\psi }\left (t \right )\right )\right )
\end{aligned} 
\label{eq:1}
\end{equation}
\indent
In the context of divergence minimization, the adversarial process between the discriminator and the generator can be represented as a two-player min-max game \citep{the-numerics-of-gans}:
\begin{equation}
\begin{aligned}
\mathop {\min }\limits_{\psi} \mathop {\max }\limits_{\theta ,\varphi} &{\mathbb{E}_{{w_t^\text{true}} \sim {p_\text{true}}}}[\log\big{(}{D_{\theta ,\varphi }}(w_t^\text{true},s_t^\text{true})\big{)}] \\
&+ {\mathbb{E}_{{w_t} \sim {\pi_{\psi}} }}[\log\big{(}1-{D_{\theta ,\varphi }}(w_t,s_t)\big{)}]\\
\end{aligned}
\label{eq:div}
\end{equation}
where ${p_\text{true}}$ is the true word distribution and $w_t^\text{true}$ is the ground truth word sampled from the true data. $s_t^\text{true}$ is the corresponding state of word $w_t^\text{true}$. In the two-player game, the discriminator maximizes the divergence between the true word distribution and the generated vocabulary distribution, whereas the generator minimizes the divergence. The adversarial training reaches Nash equilibrium when the generated vocabulary distribution ${\pi_\psi}$ approximates the estimated data distribution ${p_{\theta,\varphi}}$, i.e., $D=0.5$, and both the discriminator and the generator converge. As a result, the discriminator estimates ${p_{\theta,\varphi} }$ that approximates the true word distribution ${p_\text{true}}$, and the generator learns an optimal vocabulary distribution $\pi_\psi$ that maximizes the reward $f_{\theta,\varphi}$ from $D_{\theta,\varphi}$. \\
\indent
As mentioned before, a \emph{compact} reward function can precisely tell the contribution of each word and thus help to locate the wrong words. AIRL can learn a \emph{compact} reward function at optimality in that 1) it disentangles word-wise reward from different image-caption pairs; 2) the reward difference of two words is positively correlated to their semantic difference if AIRL can recover the true reward for each word. However, AIRL is difficult to converge to Nash equilibrium using policy gradient, requiring Hessian of the gradient vector filed being positive definite (see details in Section \ref{sec:gen}). When the adversarial training of AIRL is not convergent, apparently the true reward can not be recovered. As a result, the learned reward function is not \emph{compact}. Besides, as a GAN based method, AIRL has a sharp decision boundary for two disjoint distributions, which means the discriminator is much stronger than the generator. The consequence is a limited diversity in the generated captions (see details in Section \ref{sec:gen}), which is called mode collapse in GAN. To solve the two major problems, we explicate in the next section about how we refine the loss function to shift the generator towards Nash equilibrium and to mitigate mode collapse in the two-player game.
\section{Learning Compact Reward for Image Captioning}
\label{sec:crl}
To address the problems discussed above, we refine the loss function to \emph{1)} find a compact reward function that can reach its optimum in the adversarial training; \emph{2)} increase diversity of the generated captions. In particular, a \emph{constant term} is used to solve \emph{1)} by shifting the generator to Nash equilibrium, and a \emph{conditional term} is introduced to solve \emph{2)} by utilizing mode control techniques. Our algorithm is detailed in Algorithm 1, where $n$ is the sentence length and $N$ denotes number of iterations.\\
\indent
In the following notations, $\theta$ and $\varphi$ are the parameters of the discriminator, and $\psi$ represents the parameter of the generator. $w_t$ and $s_t$ denote the $t_{th}$ word and its corresponding hidden state vector, respectively. For better clarity, policy ${\pi_{\psi}}$ is hereinafter referred to as the generated vocabulary distribution
\subsection{Discriminator}
The objective of the discriminator is to distinguish the true caption from the generated one. At time $t$, the discriminator maximizes the divergence in Eq. (\ref{eq:div}) by
\begin{equation}
\begin{aligned}
{L_t}(\theta ,\varphi )= &-{\mathbb{E}_{w_t^{{\text{true}}}\sim{p_{{\text{true}}}}}}[\log ({D_{\theta ,\varphi }}(w_t^{{\text{true}}},s_t^{{\text{true}}}))]\\
&-{\mathbb{E}_{{w_t}\sim{\pi _\psi }}}[\log (1 - {D_{\theta ,\varphi }}({w_t},{s_t}))] \\
=&- \log {\big{(}{D_{\theta ,\varphi }}({w_t^\text{true},s_t^\text{true}})\big{)}}- \log {\big{(}1 - {D_{\theta ,\varphi }}({w_t,s_t})\big{)}}
\end{aligned}
\label{eq:dis}
\end{equation}
where $w_t^\text{true}$ is the true word and $s_t^\text{true}$ is its corresponding state. The expectation disappears since it is estimated by sampling a mini-batch from the corresponding distribution. ${D_{\theta ,\varphi }}$ is computed as in Eq. (\ref{eq:1}), where the discriminator learns the reward function ${f_{\theta,\varphi}}$ for ${D_{\theta ,\varphi }}$ and the generator estimates the vocabulary distribution $ {\pi_{\psi}}$ for ${D_{\theta ,\varphi }}$, respectively.
\begin{figure*}[t]
\centering
    \subfigure[]{ 
    \includegraphics[width=0.3\textwidth]{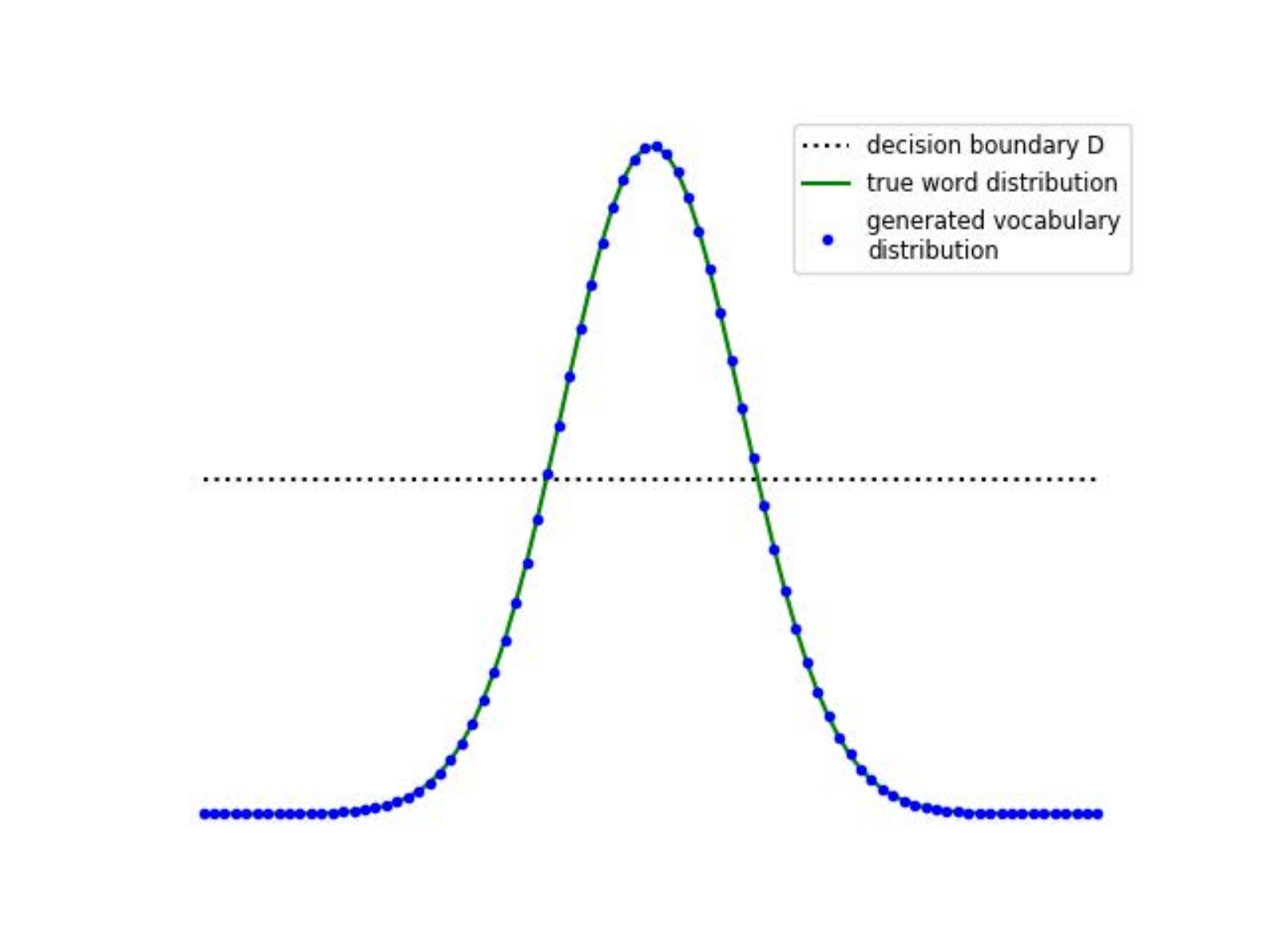}} 
    \subfigure[]{ 
    \includegraphics[width=0.3\textwidth]{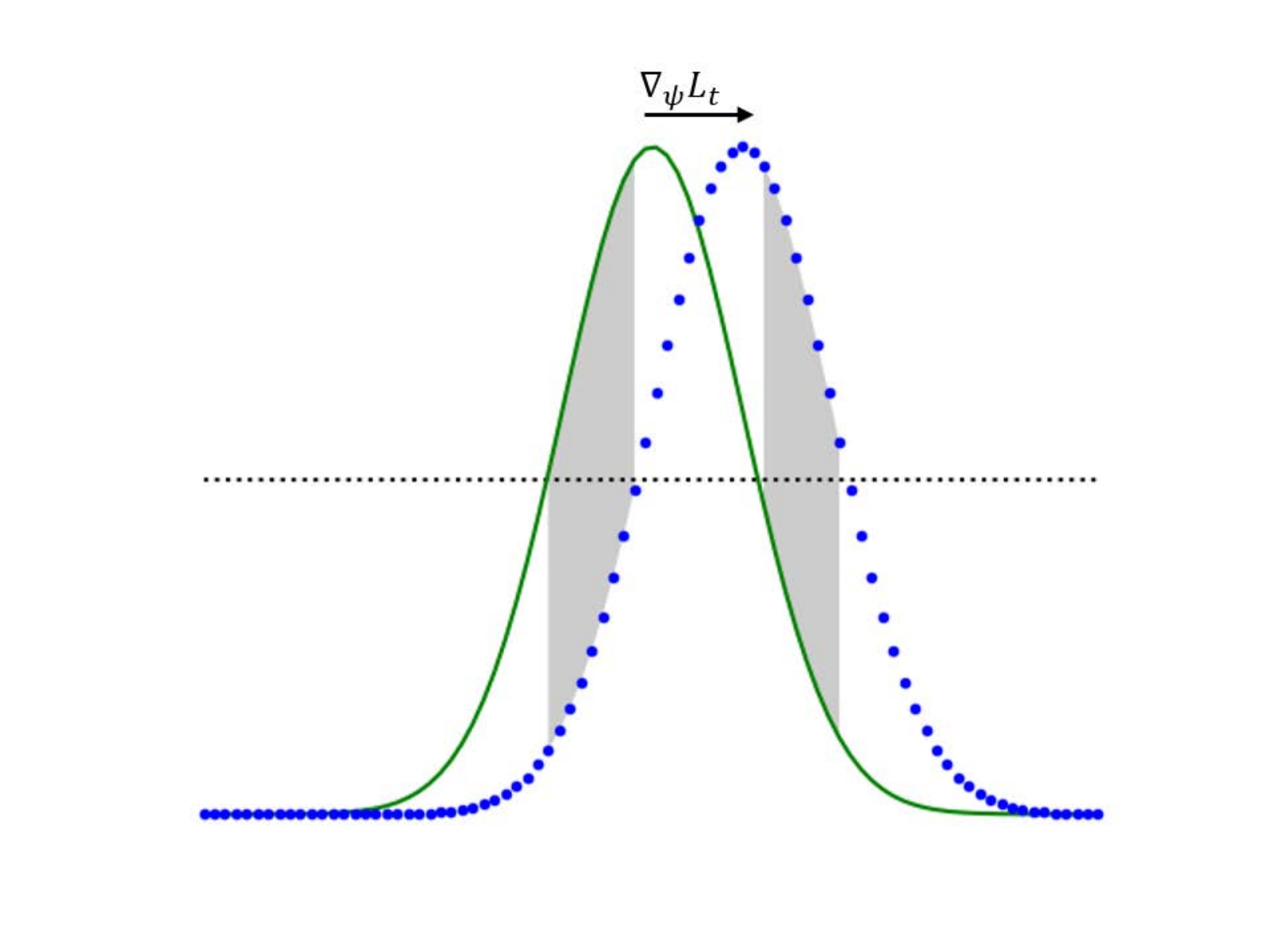}} 
    \subfigure[]{ 
    \includegraphics[width=0.3\textwidth]{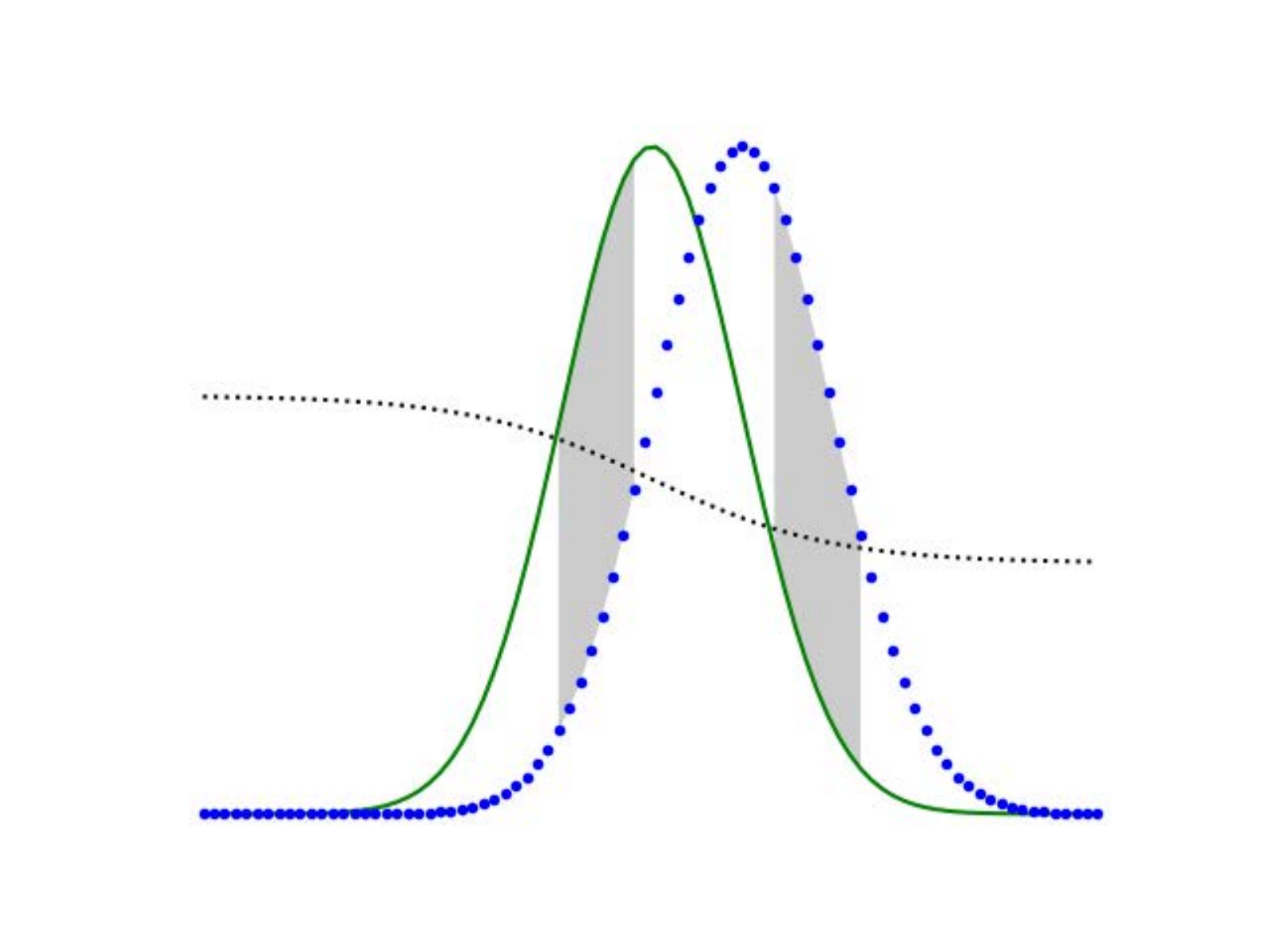}} 
  \caption{\small{Dynamics of the AIRL. Figure (a) is a recreation of Figure 1(d) in \citep{goodfellow2014generative}. \textbf{(a)} when the decision boundary $D$ (black dotted line) reaches its optimum $D=0.5$, it can not distinguish since the generated vocabulary distribution (blue dotted line) approximates the true word distribution (green line). \textbf{(b)} if the Hessian of the generator is not positive definite, gradient from the generator ${\nabla _\psi }{L_t}$ pushes it away from the equilibrium point in (a). Thus the decision boundary can discriminate between the generated vocabulary distribution and the true distribution in the gray area. \textbf{(c)} the discriminator tries to maximize the divergence between the two distribution and deviates from the optimum $D=0.5$. Such dynamics in the AIRL algorithm lead to non-convergence. Our refined algorithm shifts ${\nabla _\psi }{L_t}$ back to $0$ and thus converges to Nash Equilibrium as in (a).}}
\label{fig:conv}
\end{figure*}
\subsection{Generator}
\label{sec:gen}
Given a word $w_t$ that is sampled from the vocabulary distribution $\pi_{\psi}$, the generator maximizes ${D_{\theta ,\varphi }}({w_t,s_t})$ by
\begin{equation}
\begin{aligned}
 &{{L_t}(\psi) } =\\
 & -{\mathbb{E}_{{w_t} \sim {\pi_{\psi}} }}[\log \big{(}{D_{\theta ,\varphi }}({w_t,s_t})\big{)}- \log \big{(}1 - {D_{\theta ,\varphi }}({w_t,s_t})\big{)}]  \\ 
 &=- {\mathbb{E}_{{w_t} \sim {\pi_{\psi}} }}[{f_{\theta ,\varphi }}({w_t,s_t}) - \log ({\pi_{\psi}}(t))]  
\end{aligned}
\label{eq:gen_loss1}
\end{equation}
Using REINFORCE algorithm \citep{sutton1998reinforcement}, the gradient ${\nabla _\psi }{L_t}$ becomes:
\begin{equation}
\begin{aligned}
&{\nabla _\psi }{L_t} =-\sum\nolimits_{\pi_{\psi}}  {\big{(}{f_{\theta ,\varphi }}({w_t,s_t}) - \log ({\pi_{\psi} }(w_t,s_t))\big{)}{\nabla _\psi }{\pi_\psi}(t)   }\\
&  -\sum\nolimits_{\pi_{\psi}}{\pi_{\psi}(w_t,s_t) {\nabla _\psi }\big{(}{f_{\theta ,\varphi }}({w_t,s_t}) - \log ({\pi_{\psi}}(t) )\big{)}}  \\
=&-\sum\nolimits_{\pi_{\psi}}  {{\pi_{\psi}}(t)  {\frac{{f_{\theta ,\varphi }}({w_t,s_t}) - \log ({\pi_{\psi} }(w_t,s_t))}{{\pi_{\psi}}(t) }}{\nabla _\psi }{\pi_\psi}(t)} \\
& -\sum\nolimits_{\pi_{\psi}} {{\pi_{\psi}}(t) {{\nabla _\psi }\big{(}{f_{\theta ,\varphi }}({w_t,s_t}) - \log ({\pi_{\psi}}(t) )\big{)}}}  \\
=&- {\frac{{f_{\theta ,\varphi }}({w_t,s_t}) - \log ({\pi_{\psi} }(t))}{{\pi_{\psi}}(t) }}{\nabla _\psi }{\pi_\psi}(t)  \\
&- {\nabla _\psi }\big{(}{f_{\theta ,\varphi }}({w_t,s_t}) - \log ({\pi_{\psi}}(t))\big{)}\\
=&  - \frac{{f_{\theta ,\varphi }}({w_t},{s_t}) - \log ({\pi_{\psi}}(t) ) - 1}{{\pi_{\psi}}(t) }{\nabla _\psi }{\pi_\psi}(t)
\end{aligned}
\label{eq:reinforce}
\end{equation}
When the decision boundary $D$ reaches its optimum ($D=0.5,f_{\theta ,\varphi }({w_t},{s_t})=\log ({\pi_\psi}(t))$) as in Figure \ref{fig:conv}(a), the generator can only converge when ${\nabla _\psi }{\pi_\psi} =0$ in Eq. (\ref{eq:reinforce}), requiring Hessian of the gradient vector filed being positive definite \citep{the-numerics-of-gans}. Otherwise, even if the generator has learned the true word distribution ($\log ({\pi_\psi}(t))=f_{\theta ,\varphi }({w_t},{s_t})=p_\text{true}))$, the non-zero gradient ${\nabla _\psi }{L_t}$ from itself still pushes it away from the true word distribution. Thus the decision boundary can discriminate between the generated vocabulary distribution and the true word distribution in the gray area of Figure \ref{fig:conv}(b). Then the discriminator tries to maximize the divergence between the two distributions and deviates from the optimum $D=0.5$, which leads to the results in Figure \ref{fig:conv}(c). Such dynamics cause non-convergence in the adversarial training. If the generator converges at $\log ({\pi_\psi}(t))=f_{\theta ,\varphi }({w_t},{s_t})$ instead of ${\nabla _\psi }{\pi_\psi} =0$ in Eq. (\ref{eq:reinforce}), then its gradient ${\nabla _\psi }{L_t}$ becomes $0$ at $D=0.5$ and the Nash equilibrium in Figure \ref{fig:conv}(a) can be maintained. Therefore, we introduce a \emph{constant term} in the expectation in Eq. (\ref{eq:gen_loss1}) 
\begin{equation}
{{L_t}(\psi) } =- {\mathbb{E}_{{w_t} \sim {\pi_{\psi}}}}[{f_{\theta ,\varphi }}({w_t,s_t}) - \log ({\pi_{\psi}}(t) )+1]
\label{eq:lt}
\end{equation}
Thus, according to Eq. (\ref{eq:reinforce}), we have
\begin{equation}
 {\nabla _\psi }{L_t}=  - \frac{1}{\pi_{\psi} }({f_{\theta ,\varphi }}({w_t},{s_t}) - \log ({\pi_{\psi}}(t)  )){\nabla _\psi }{\pi_{\psi}}(t) 
 \label{eq:lt2}
\end{equation}
such that the generator converges at the equilibrium point where $f_{\theta ,\varphi }({w_t},{s_t})=\log ({\pi_{\psi}}(t) )$, i.e., Figure \ref{fig:conv}(a). It is noted that the \emph{constant term} can also be regarded as \emph{baseline} in REINFORCE, except it is utilized to centralize the stationary point instead of reducing variance of the estimation.\\
\indent 
The expectation has been removed from the gradient ${\nabla _\psi }{L_t}$ using REINFORCE. Thus the expectation in the loss function disappears by taking the integral of its gradient w.r.t. $\psi$, for which we have
\begin{equation}
\begin{aligned}
&{L_t}(\psi ) =  - {\mathbb{E}_{{w_t}\sim{\pi _\psi }}}[{f_{\theta ,\varphi }}({w_t},{s_t}) - \log ({\pi _\psi }(t)) +1] \\
&=\int\limits_\psi  {{\nabla _\psi }{L_t}} {\text{d}}\psi \\
&= -\int\limits_{\psi }{ \frac{{f_{\theta ,\varphi }}({w_t},{s_t}) - \log ({\pi_{\psi}}(t) )}{{\pi_{\psi}}(t) }{\nabla _\psi }{\pi_{\psi}}(t){\text{d}\psi }}\\
&= -{\big{(}{f_{\theta ,\varphi }}({w_t},{s_t})- \log ({\pi_{\psi}}(t))\big{)}\log ({\pi_{\psi}}(t) ) }
\end{aligned}
\end{equation}
\begin{figure*}[t]
\centering
    \subfigure[]{ 
     \label{fig:cond1}
    \includegraphics[width=0.4\textwidth]{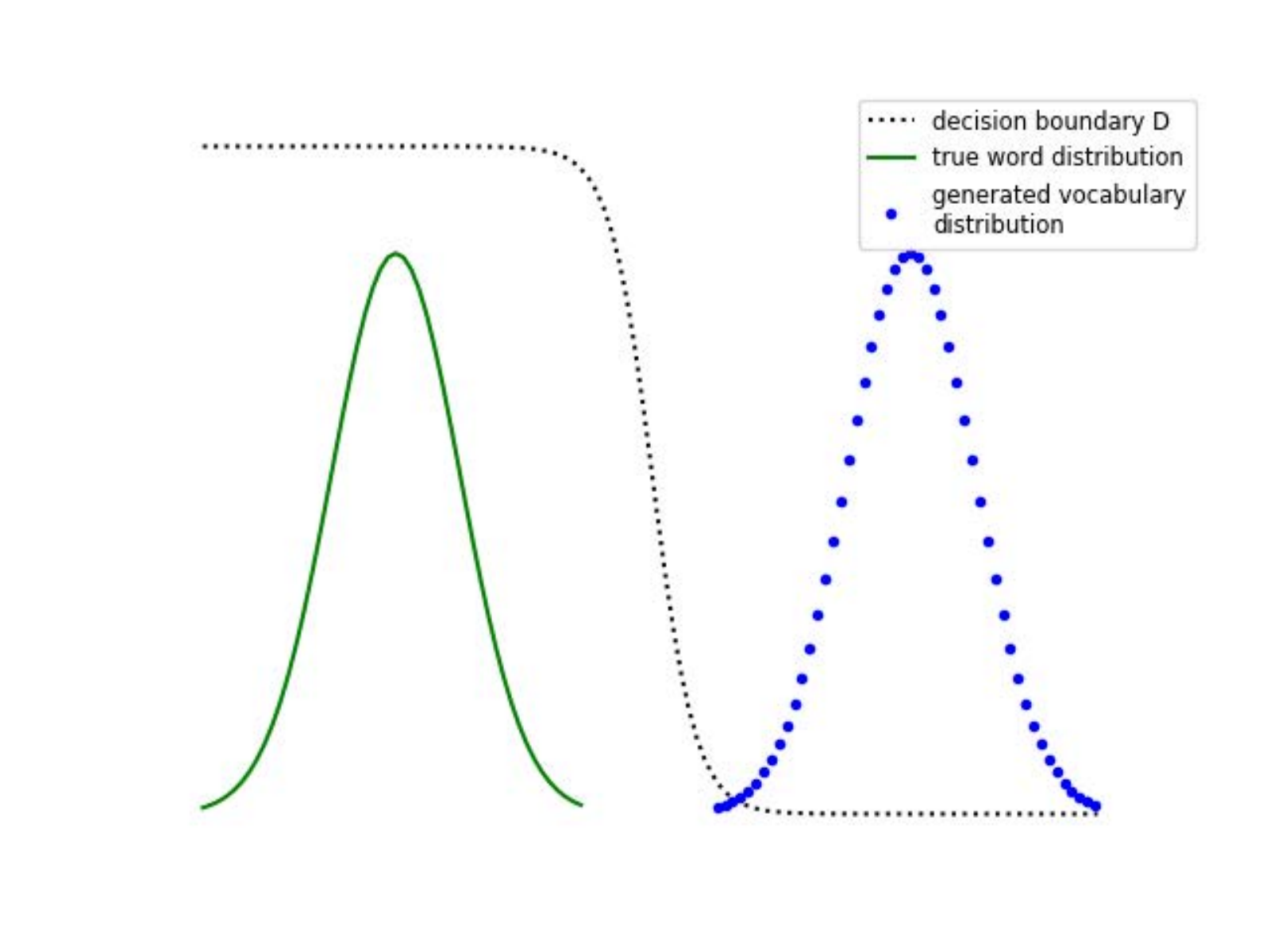}}    
    \subfigure[]{ 
     \label{fig:cond2}
    \includegraphics[width=0.4\textwidth]{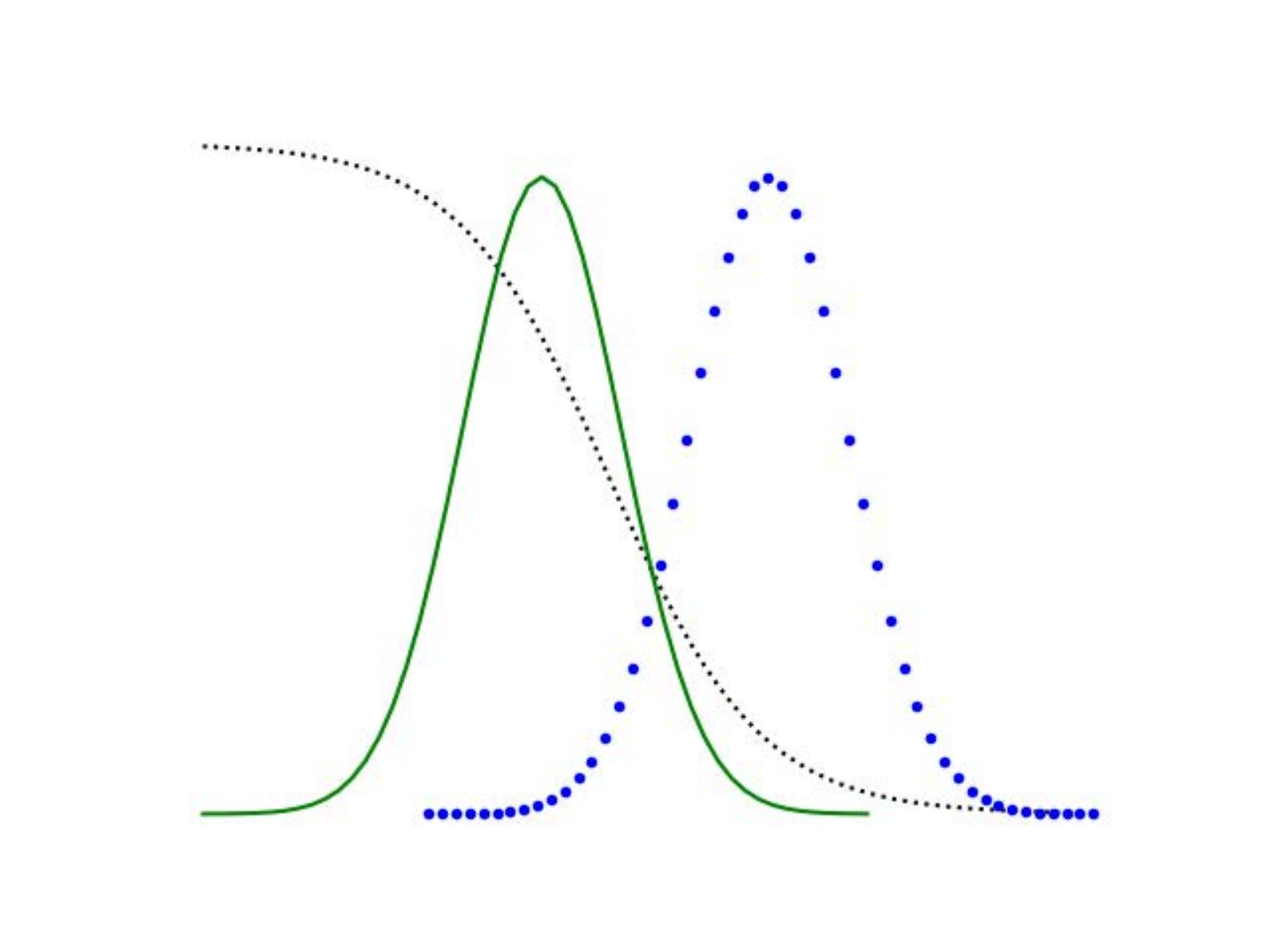}}    
  \caption{\small{Illustration of the effects of the conditional term. a) The gradients of the discriminator are zero most of the time for differentiating between two disjoint distributions, which occurs when the generator hasn't approximated the true word distribution. b) The conditional term smooths the decision boundary by mixing up the true data distribution and the generate data distribution.}}
\label{fig:cond}
\end{figure*}
\indent
In practice, the discriminator is usually easier to converge than the generator. If the discriminator converges too early, the generated vocabulary distribution hasn't approximated the true word distribution, which makes them two disjoint distributions. The gradients of $D$ are thus zero almost everywhere (see Figure \ref{fig:cond1}) \citep{peng2018variational}, causing limited diversity of the generated captions. The problem is called mode collapse, meaning that the generator produces a single or limited modes. If the generated vocabulary distribution has some overlap with the true word distribution, then the discriminator can not easily differentiate between them, which makes the decision boundary $D$ more smooth. Therefore, we introduce ground truth data into the generator as a \emph{conditional term} \citep{MirzaO14}:
\begin{equation}
\begin{aligned}
&{{L_t}(\psi) } =- {\mathbb{E}_{{w_t} \sim {\pi_{\psi}}}}[{f_{\theta ,\varphi }}({w_t,s_t})- \log ({\pi_{\psi}}(t) )+1]\\
&- {\mathbb{E}_{w_t^\text{true} \sim \pi _\psi ^\text{true}}}[{f_{\theta ,\varphi }}(w_t^\text{true},s_t^\text{true}) - \log ({\pi _\psi ^\text{true}}(t)) + 1] \\
&=  -{\big{(}{f_{\theta ,\varphi }}({w_t},{s_t})- \log ({\pi_{\psi}}(t)  )\big{)}\log ({\pi_{\psi}}(t) ) }\\
&- \big{(}{f_{\theta ,\varphi }}(w_t^\text{true},s_t^\text{true}) - \log ({\pi _\psi ^\text{true}}(t))\big{)} \log ({\pi _\psi ^\text{true}}(t))
\end{aligned}
\label{eq:gen}
\end{equation}
where $\pi_{\psi}^\text{true}$ is the approximated probability of the true word in the generator, and $ {\mathbb{E}_{w_t^\text{true} \sim \pi _\psi ^\text{true}}}[\cdot] $ is the \emph{conditional term}. To give a rough idea, the conditional term smooths the decision boundary by increase the overlap between the true word distributions and the generated vocabulary distribution (see Figure \ref{fig:cond2}) by introducing ground truth data. It strengthens the generator in the following way: 1) When $D_\text{true} > D_\text{gen}$ at time $t$, the gradient of the true word becomes larger than that of the generated one (${\nabla _{\pi_{\psi}^\text{true}}}{L_t} > {\nabla _{\pi_{\psi}}}{L_t}$), and thus the generator further increases the probability of the true word (${\pi_{\psi}^\text{true}}$). 2) Otherwise (i.e., $D_\text{true} < D_\text{gen}$), the generator prefers sampling its self-generated word to confuse the discriminator. By picking the one from the two distributions that can fool the discriminator, the generated vocabulary distribution has more overlap with the true word distribution, which smooths the decision boundary as in Figure \ref{fig:cond2} and thus helps $D$ maintain informative gradient during the adversarial training. The coefficient of $log ({\pi _\psi ^\text{true}}(t))$ is symmetrical to the coefficient of $\log ({\pi_{\psi}}(t))$ and is updated adaptively in the training process. Note that adding the conditional term does not change the model's convergence to Nash equilibrium since $\pi_{\psi}=\pi_{\psi}^\text{true}$ at the equilibrium.
\begin{table}[h]
		\centering
		\small
		\caption{\small{Formulas of different loss functions.}}
		\scalebox{0.8}[0.8]{
\begin{tabular}{cc}\toprule
	   Method  &Loss Function\cr  \midrule
	   MLE &$- \sum\limits_{t = 1}^n {\log (\pi _t^{{\text{true}}})}  - \sum\limits_{t = 1}^n {\sum\limits_{{\pi _t} \ne \pi _t^{{\text{true}}}} {\log (1 - {\pi _t})} } $ \cr  \midrule
	   RL &$- r\sum\limits_{t = 1}^n {\log({\pi _t})}$\cr \midrule
	   GAN (generator) &$ - {D_{{\text{gen}}}}\sum\limits_{t = 1}^n {\log ({\pi _t})}$\cr \midrule
	   rAIRL (generator) &$ - \sum\limits_{t = 1}^n {{\sigma ^{ - 1}}(D_t^{{\text{gen}}})\log ({\pi _t}) - {\sigma ^{ - 1}}(D_t^{{\text{true}}})\log( \pi _t^{{\text{true}}})}$ \cr
		\bottomrule	
		\end{tabular}	
		}
		\label{tab:loss}
\end{table}
\subsection{Discussion on Loss Functions}
We compare the formula of the proposed loss function with existing methods in Table \ref{tab:loss}, including MLE, RL and GAN. $n$ is the length of a sentence. $r$ is the handcrafted metric, such as BLEU, CIDEr and SPICE. $\pi_t$ is the probability of the $t_\text{th}$ generated word, and ${\pi_t}^\text{true}$ is the probability of the $t_\text{th}$ true word. The loss functions are rewritten using similar symbols for direct comparison. MLE maximizes the probability of the true data ${\pi _t^{{\text{true}}}}$ whist RL and GAN optimize the reward by sampling from ${\pi _t}$. GAN is different from RL in that its reward is learned from the discriminator adversarially instead of being predefined. GAN is capable of mimicking human-written captions by adversarial learning, but the estimated reward function $D_{{\text{gen}}}$ of a full trajectory can be explained by multiple optimal policies and thus is too ambiguous. The proposed rAIRL further disentangles the reward into $D_t^{{\text{gen}}}$ at each time step $t$, as well as incorporating the true data for better diversity. From the perspective of loss functions, rAIRL can be regarded as an integration of GAN and the first term of MLE using coefficients ${{\sigma ^{ - 1}}(D_t^{{\text{gen}}})}$ and ${{\sigma ^{ - 1}}(D_t^{{\text{true}}})}$.
\section{Experiments}
In the experiments, we validate the effectiveness of the proposed algorithm by answering two questions: 1) Is the caption evaluator (i.e., discriminator) capable of learning compact reward?  2) Driven by the learned reward, is the caption generator effective to produce qualitative captions?\\
\indent
To answer 1), we first tested the compactness of the learned reward by observing performance of the collected top-$k$ captions. Then we explored the correlation between the learned reward and the human evaluation results. To answer 2), we built our algorithm on existing learning methods and compared their performance on conventional evaluation metrics. For a comprehensive evaluation, we also evaluated the quality of the generated caption on its content, diversity and grammar. Besides, ablation experiments were conducted to demonstrate the importance of each component of our algorithm.
\begin{figure*}[!t]
\centering  
    \includegraphics[width=0.95\textwidth]{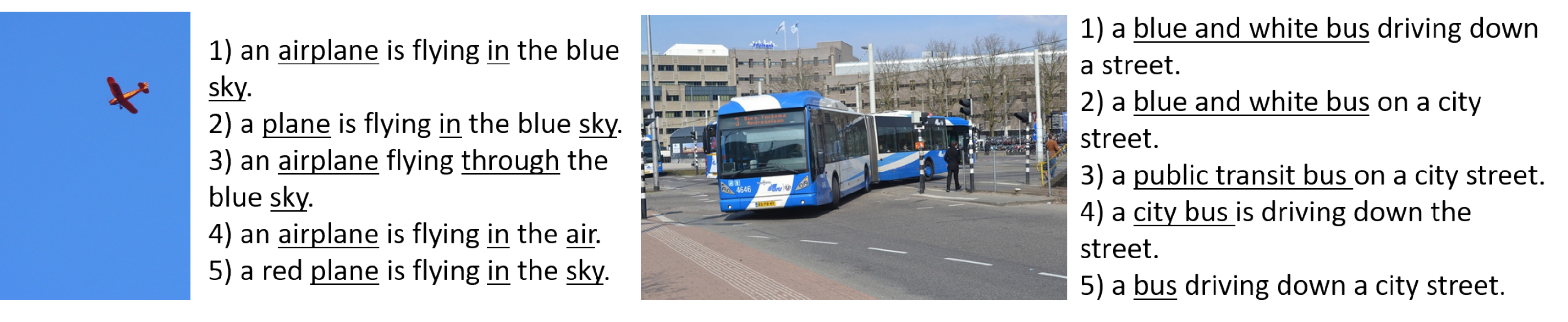} 
  \caption{\small{Examples of the top-$5$ generated captions of rAIRL. Driven by the compact reward function, the generator describes a given image with semantically similar words.}}
   \label{fig:compact}
\end{figure*}
\subsection{Implementation Details}
We conducted experiments on the well-known benchmark datasets MS COCO \citep{capeval2015} and Flickr30K \cite{Young2014From}, which have 123,287 and 31,783 labeled images, respectively, and each image has at least 5 human annotated captions as reference. We use the public available split \cite{Karpathy2015Deep} for Flickr30K. To assess the robustness of our algorithm, we use two splits of the COCO dataset: standard split \citep{Karpathy2015Deep} which is created by randomly picking test images, and robust split \citep{Lu2018Neural} which is organized to maximize difference of the co-occurrence distribution between the training and test set. The robust split is recently proposed and is more challenging due to its distribution difference between the training and test set. The standard split has $113,287/5000/5000$ train/val/test images and the robust split has $110,234/$ $3915/9138$ train/val/test images. \\
\indent
We implement our algorithm using Adam optimizer \citep{kingma2014adam} with fixed learning rate $10^{-5}$. Our vocabulary size is $10,000/7065$ for MS COCO and Flickr30K, respectively, including a special start sign \textless BOS\textgreater and an end sign \textless EOS\textgreater. In the generator, the number of hidden nodes of every layer is $512$. For simplicity, the discriminator has the same model structure as the generator except having one additional layer for estimating $h_{\varphi}$. For fair comparison, all the methods in ML (Up-Down), RL(Up-Down), GAN(Up-Down), AIRL(Up-Down) and rAIRL(Up-Down) were produced using the same image features and model structure (Up-Down) in \citep{peter2017bot}. Specifically, RL(Up-Down) adopts the self-critical loss in \citep{Rennie2016Self}. GAN(Up-Down) uses the adversarial loss in \citep{DaiL17Con} that learns sentence-level reward. AIRL(Up-Down) is the standard adversarial inverse reinforcement learning method in \citep{Fu2018LearningRobust}, and rAIRL(Up-Down) is the proposed method. Note that our scores of MLE(Up-Down) are lower on the standard split but higher on the robust split than \citep{peter2017bot} because we used fixed number of the bounding box (i.e., $36$) for simplicity, and the hyperparameters were tuned to adapt to both splits and thus are not exactly the same with \citep{peter2017bot}.
\subsection{Performance of the Recovered Reward}
\label{sec:reward}
In this section, we evaluated the learned reward function on three aspects: compactness, correlation with human evaluation and performance on diagnosing captions. Compactness shows accuracy of the disentangled reward in measuring word semantic. It is evaluated by computing the correlation between the reward differences and semantic differences after replacing specific words in the caption. Correlation with human evaluation indicates how well the caption evaluator correlates with human judgments. It is based on human scores collected by \citet{Aditya2017}. Diagnosing captions using the learned reward function can help improving captions, whose performance is evaluated by the relative improvement after diagnosing and re-written.\\
\begin{table}[h]
		\centering
		\caption{\small{Correlation between the reward differences and semantic differences by replacing a given word with a similar word (RP\underline{~~}S) and a distinct word (RP\underline{~~}D), respectively.\\}}
		\begin{tabular}{ccccc}\toprule
		 \multirow{2}{*}{Method } &\multicolumn{2}{c}{Standard Split} &\multicolumn{2}{c}{Robust Split} \cr \cmidrule(r){2-3} \cmidrule{4-5}
		&RP\underline{~~}S &RP\underline{~~}D &RP\underline{~~}S &RP\underline{~~}D  \cr \midrule		 
		RL(Up-Down) &0.03 &0.01 &-0.10 &0.00 \cr
		GAN(Up-Down)  &0.07 &0.21 &0.04 &0.18 \cr
		AIRL(Up-Down) &0.15 &0.11 &0.30 &0.20 \cr
		rAIRL(Up-Down) &\textbf{0.54} &\textbf{0.30} &\textbf{0.51} &\textbf{0.31}  \cr				
		\bottomrule
		\end{tabular}	
		\label{tab:compact}	
\end{table} 
\begin{table*}[!t]
		\centering
		\caption{\small{Sentence-level correlation with human evaluation. All p-value (not shown) are less than 0.001.\\}}
		\begin{tabular}{ccccccc}\toprule
		 \multirow{2}{*}{Method } &\multicolumn{3}{c}{Correctness} &\multicolumn{3}{c}{Throughness} \cr \cmidrule(r){2-4} \cmidrule{5-7}
		&Peason &Spearman &Kendall  &Peason &Spearman &Kendall  \cr \midrule
		BLEU1 &0.19 &0.27 &0.19 &0.20 &0.28 &0.20\cr
		BLEU4 &0.33 &0.30 &0.22 &0.32 &0.31 &0.22\cr
		CIDEr &0.40 &0.45 &0.37 &0.41 &0.45 &0.36\cr
		SPICE &0.44 &0.45  &0.39 &0.45 &0.46 &0.38\cr			
		GAN(Up-Down)  &0.12 &0.11 &0.15 &0.12 &0.11 &0.15\cr
		AIRL(Up-Down) &0.04 &0.06 &0.08 &0.05 &0.06 &0.07\cr
		rAIRL(Up-Down) &0.43 &0.40 &0.35 &0.40 &0.37 &0.34 \cr				
		rAIRL+BLEU1(Up-Down) &0.44 &0.41 &0.35 &0.41 &0.39 &0.34\cr
		rAIRL+BLEU4(Up-Down) &0.45 &0.43 &0.36 &0.42 &0.42 &0.35\cr
		rAIRL+CIDEr(Up-Down) &0.43 &0.45 &0.38 &0.42 &0.46 &0.37\cr
		rAIRL+SPICE(Up-Down) &\textbf{0.47} &\textbf{0.46} &\textbf{0.41} &\textbf{0.46} &\textbf{0.47} &\textbf{0.39} \cr	
		\bottomrule
		\end{tabular}	
		\label{tab:corr}	
\end{table*} 
\begin{table*}[!t]
		\centering
		\caption{\small{Results of rewriting caption from the located position by rAIRL on MS COCO standard split. Beside each score we report its improvement relative to rewriting from a random position.\\}}
		\scalebox{.8}[.8]{
		\begin{tabular}{ccccccccccccccc}\toprule
		\multirow{2}{*}{Source Caption}  &\multicolumn{2}{c}{BLEU1} &\multicolumn{2}{c}{BLEU2}  &\multicolumn{2}{c}{BLEU3}  &\multicolumn{2}{c}{BLEU4} &\multicolumn{2}{c}{ROUGE\_L} &\multicolumn{2}{c}{CIDEr} &\multicolumn{2}{c}{SPICE}\cr  \cmidrule(r){2-3}  \cmidrule(r){4-5}  \cmidrule(r){6-7}  \cmidrule(r){8-9}  \cmidrule(r){10-11}  \cmidrule(r){12-13}  \cmidrule(r){14-15}
		&score &$\Delta$ &score &$\Delta $ &score &$\Delta $ &score &$\Delta $ &score &$\Delta $ &score &$\Delta $ &score &$\Delta $ \cr \midrule
       MLE(Up-Down)     &73.3 &(+0.0)	&57.1 &(-0.2)	 &43.4 &(+0.1)	&32.6 &(+0.5)	&51.4 &(+0.1)	&108.5 &(+2.0)	 &20.6 &(+0.1)\cr       
       RL(Up-Down)       &72.9 &(-0.1)	&56.7 &(+0.3)  &42.4 &(+0.5)	&31.1 &(+0.4)	&50.8 &(+0.2)	&104.0 &(-0.5)	&20.1 &(+0.0) \cr
       GAN(Up-Down)    &72.7 &(+1.1)	&56.3 &(+1.2)	 &42.0 &(+1.0)	&30.9 &(+0.8)	&50.6 &(+0.4)	&103.0 &(+2.3)	&20.0 &(+0.4)\cr
       AIRL(Up-Down)    &72.6 &(+1.0)	&56.1 &(+1.3)	 &41.8 &(+1.2)	&30.5 &(+0.9)	&50.6 &(+0.8)	&102.6 &(+4.1)	&19.9 &(+0.8)\cr
		\bottomrule
		\end{tabular}			
		}
		\label{tab:diag}	
\end{table*}
\paragraph{\textbf{Compactness.}} Compactness means that the reward values should be close for similar words and different for distinct words. For example, \emph{kid} can also be referred to as \emph{little boy} or \emph{little girl}, and thus their reward values should be close to each other in the discriminator. To see the correlation between the reward differences and semantic differences, we replace a given word $w_t$ in the generated caption with a similar word $w_\text{similar}$ and a distinct word $w_\text{distinct}$, respectively. Specifically, in a generated caption, the first word that belongs to the COCO $80$ class\footnote{\url{https://github.com/jiasenlu/NeuralBabyTalk/blob/master/data/coco/coco_class_name.txt}} \citep{Lu2018Neural} is replaced. A sentence is discarded if no word can be replaced. The words within the same class are considered to be similar  (such as \emph{bike} and \emph{bicycle}), and the words that belong to difference classes are distinct (such as \emph{man} and \emph{bike}). For RL, since it maximizes a handcrafted reward (SPICE \citep{anderson16spice} in our experiment) instead of learning a reward function, the reward difference is the variation of SPICE before and after replacement. For reward-learning methods, the reward difference is the variation of the learned reward. The semantic difference is the Euclidean distance between the Glove embedding vectors of two words \citep{pennington2014glove}. The results are reported in Table \ref{tab:compact}. Higher correlation indicates better compactness. RL serves as a baseline in that the handcrafted reward SPICE compares $n$-gram overlapping without considering the semantic difference. The reward differences of rAIRL correlate the best with the semantic differences for both similar words and distinct words, proving the compactness of the learned reward. It's also noted that due to the reward ambiguity problem, the reward differences of GAN poorly correlate with the semantic differences, especially for similar words.\\
\indent
Figure \ref{fig:compact} shows the top-$5$ generated captions of an given image. Driven by the learned compact reward function, the top-$5$ captions of an image have the same format whilst some words are replaced with their synonyms. For example, the model uses \emph{blue and white bus}, \emph{city bus} and \emph{public transit bus} to describe the bus in the second picture, while the format of its generated sentence remains the same.
\paragraph{\textbf{Correlation with human evaluation.}}  As a caption evaluator, the discriminator learns $g_{\theta}$ that recovers the true reward up to a constant at optimality \citep{Fu2018LearningRobust}. To explore the correlation between the recovered reward and the human evaluation scores, we used the human scores in the COMPOSITE\footnote{\url{https://imagesdg.wordpress.com/image-to-scene-description-graph/}} dataset \citep{Aditya2017}, whose images are subsets from Flickr8k, Flickr30k and MS COCO. The descriptions from this dataset are either ground truth captions or generated sentences by \citep{Aditya2017,johnson2015image}. In the human evaluation process, the AMT worker was asked to give a score at range of 1-5 to evaluate the correctness and throughness of each sentence. Captions with length exceeding $20$ were discarded, resulting a total of $11,657$ sentences. Full results of the correlation is shown in Table \ref{tab:corr}. The correlation is evaluated using Pearson $p$, Kendall's $\tau$ and Spearman's $r$ correlation coefficients. \\
\indent
In Table \ref{tab:corr}, the reward of AIRL/rAIRL is the sum of the word-wise reward $g_{\theta}$, and the reward of rAIRL+SPICE is a linear combination of $g_{\theta}$ and the SPICE score. Among the reward-learning methods, AIRL poorly correlates with human, whereas the proposed rAIRL improves AIRL on all the correlation metrics, especially on the Pearson correlation (from $0.04$ to $0.43$). Furthermore, a simple combination of SPICE and the recovered reward leads to an increased correlation with the human scores, which proves the capacity of the discriminator as a caption evaluator. We also found that conventional metrics, especially BLEU, do not correlate well with human evaluation in terms of linear correlation. Therefore, in the experiments of evaluating the captions in the next section, we directly adopt human studies as the evaluation method, along with other objective evaluation metrics that have proven to correlate well with human, including SPICE \citep{anderson16spice}, CHAIR$_s$ and CHAIR$_i$ \citep{rohrbach2018object}. Results on the conventional metrics are also reported for comparison with existing methods, but they are not addressed.
\begin{figure*}[!t]
\centering
    \includegraphics[width=0.95\textwidth]{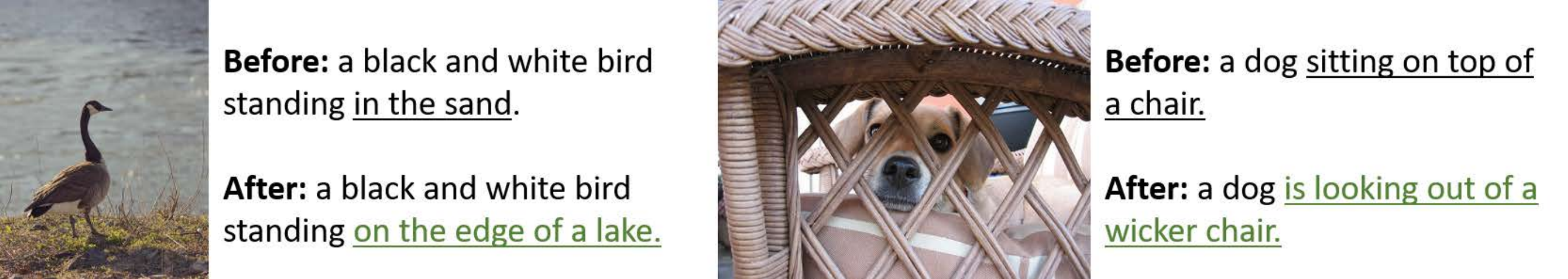}
  \caption{\small{Examples showing the generated captions from AIRL before and after re-written. }}
  \label{fig:diagnoise}
\end{figure*}
\begin{table*}[!t]
		\centering
		\caption{\small{Comparison with existing methods on the handcrafted evaluation metrics.\\}}
		\scalebox{0.9}[0.9]{
		\begin{tabular}{ccccccccccc}\toprule
		\multirow{2}{*}{Learning Method} &\multirow{2}{*}{Model} &\multicolumn{3}{c}{Standard Split} &\multicolumn{3}{c}{Robust Split} &\multicolumn{3}{c}{Flickr30K} \cr \cmidrule(r){3-5} \cmidrule(r){6-8} \cmidrule{9-11}
        &&BLEU4 &CIDEr &SPICE &BLEU4 &CIDEr &SPICE &BLEU4 &CIDEr &SPICE \cr \midrule
       \multirow{4}{*}{MLE } 
       &Att2in &31.3 &101.3 &- &31.5 &90.6 &17.7 &- &- &-\cr
       &NBT &34.7  &107.2 &20.1 &31.7 &94.1 &18.3 &27.1 &57.5 &15.6\cr     
       &Up-Down &\textbf{36.2} &\textbf{113.5} &20.3 &\textbf{31.6} &92.0 &18.1 &- &- &-\cr
       &rAIRL+MLE(Up-Down) &34.6 &112.9 &\textbf{20.7} &31.1 &\textbf{96.8} &\textbf{19.1}  &\textbf{29.2} &\textbf{58.9} &\textbf{15.7}\cr  \midrule
       \multirow{4}{*}{RL } 
       &GAN$_2$ &- &111.1 &- &- &- &- &- &- &-\cr
       &Att2in &33.3 &111.4 &- &- &- &- &- &- &- \cr      
       &Up-Down &\textbf{36.3} &\textbf{120.1} &\textbf{21.4} &- &- &- &- &-  &-\cr      
       &rAIRL+RL(Up-Down) &35.0 &115.7 &21.3 &30.8 &97.9 &19.7 &28.4 &57.5 &15.6\cr  \midrule
        \multirow{3}{*}{GAN } 
        &G-GAN &20.7&79.5 &18.2 &- &- &- &8.8 &20.2 &8.7 \cr
        &GAN$_3$ &- &97.5 &- &- &- &- &- &- &-\cr
      &rAIRL(Up-Down) &\textbf{33.8} &\textbf{110.2} &\textbf{20.4} &\textbf{30.2} &\textbf{93.7} &\textbf{18.7} &\textbf{28.7} &\textbf{55.6} &\textbf{15.6}\cr           
		\bottomrule
		\end{tabular}	
		}
		\label{tab:all}	
\end{table*}
\paragraph{\textbf{Diagnose and improve captions.}} Since the proposed rAIRL learns a word-wise reward, it's also applicable to diagnose a given caption by finding the wrong word (e.g., the word whose reward decreases sharply compared with that of the previous word) and rewriting the caption to improve its quality. For example, improving \emph{a man is playing soccer} to \emph{a man and a kid are playing soccer} can be done by rewriting the caption from \emph{is} (see Figure \ref{fig:diagnoise} for more examples). Therefore, we choose to rewrite a given caption (source caption) from the word whose reward has a decrease rate larger than $50\%$. However, we found that even rewriting the source caption from a random position using rAIRL can also improve the evaluation scores. Thus, rewriting from a random position is selected as the baseline to compare with rewriting from the located position. Table \ref{tab:diag} shows results of rewriting from the located position, where the source captions are given by MLE, RL, GAN and AIRL. Beside each score we report its improvement relative to rewriting from a random position, whose values are mostly positive. This demonstrates that the proposed rAIRL can diagnose the caption at a word level, and further improves the caption quality by rewriting from the located position.
\begin{table*}[!t]
		\centering
		\caption{\small{Evaluation scores on generated captions. The best score is in bold font and the second best score is underlined. SPICE is the handcrafted evaluation metric. CHAIR$_s$ and CHAIR$_i$ represent the object hallucination ratio at sentence level and instance level, respectively. \textbf{Human} indicates human evaluation. \\}}		
		\begin{tabular}{ccccccccc}\toprule
		 \multirow{2}{*}{Method } &\multicolumn{4}{c}{Standard Split} &\multicolumn{4}{c}{Robust Split} \cr \cmidrule(r){2-5} \cmidrule{6-9}
		 &SPICE  &CHAIR$_s$ &CHAIR$_i$ &Human  &SPICE &CHAIR$_s$ &CHAIR$_i$  &Human  \cr \midrule
       MLE(Up-Down)    &19.0  &8.3  &\underline{6.0}  &{16.1}   &\underline{18.6}   &19.1 &16.9 &{18.0}  \cr       
       RL(Up-Down)   &\textbf{20.7} &11.4 &8.5  &{8.7}    &18.1  &25.2 &20.4 &{6.4}   \cr
       GAN(Up-Down) &18.3  &\underline{7.6}  &6.4     &\underline{19.9} &16.8  &\underline{17.3} &\underline{15.2} &\underline{20.2} \cr
       AIRL(Up-Down) &17.3  &12.7 &10.3   &14.0    &16.7  &22.7 &18.5 &{14.8}   \cr
       rAIRL(Up-Down) &\underline{20.4} &\textbf{7.2}   &\textbf{5.5}  &\textbf{41.3}   &\textbf{18.7}  &\textbf{17.1} &\textbf{14.3} &\textbf{40.6}    \cr
		\bottomrule
		\end{tabular}			
		\label{tab:cd}	
\end{table*}
\begin{figure*}[!t]
\centering
     \subfigure[]{ 
    \includegraphics[width=0.3\textwidth]{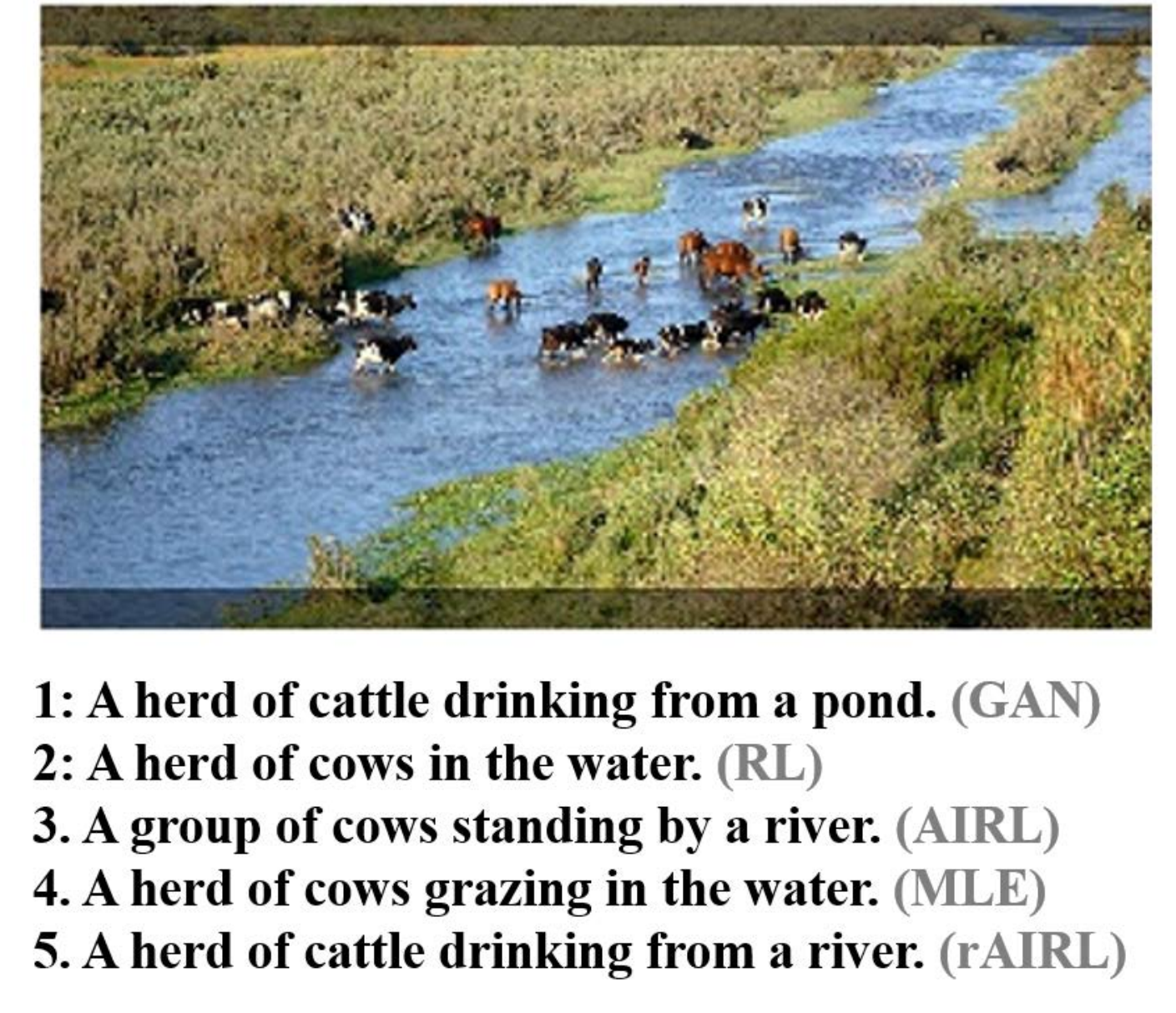}} 
    \subfigure[]{ 
    \includegraphics[width=0.4\textwidth]{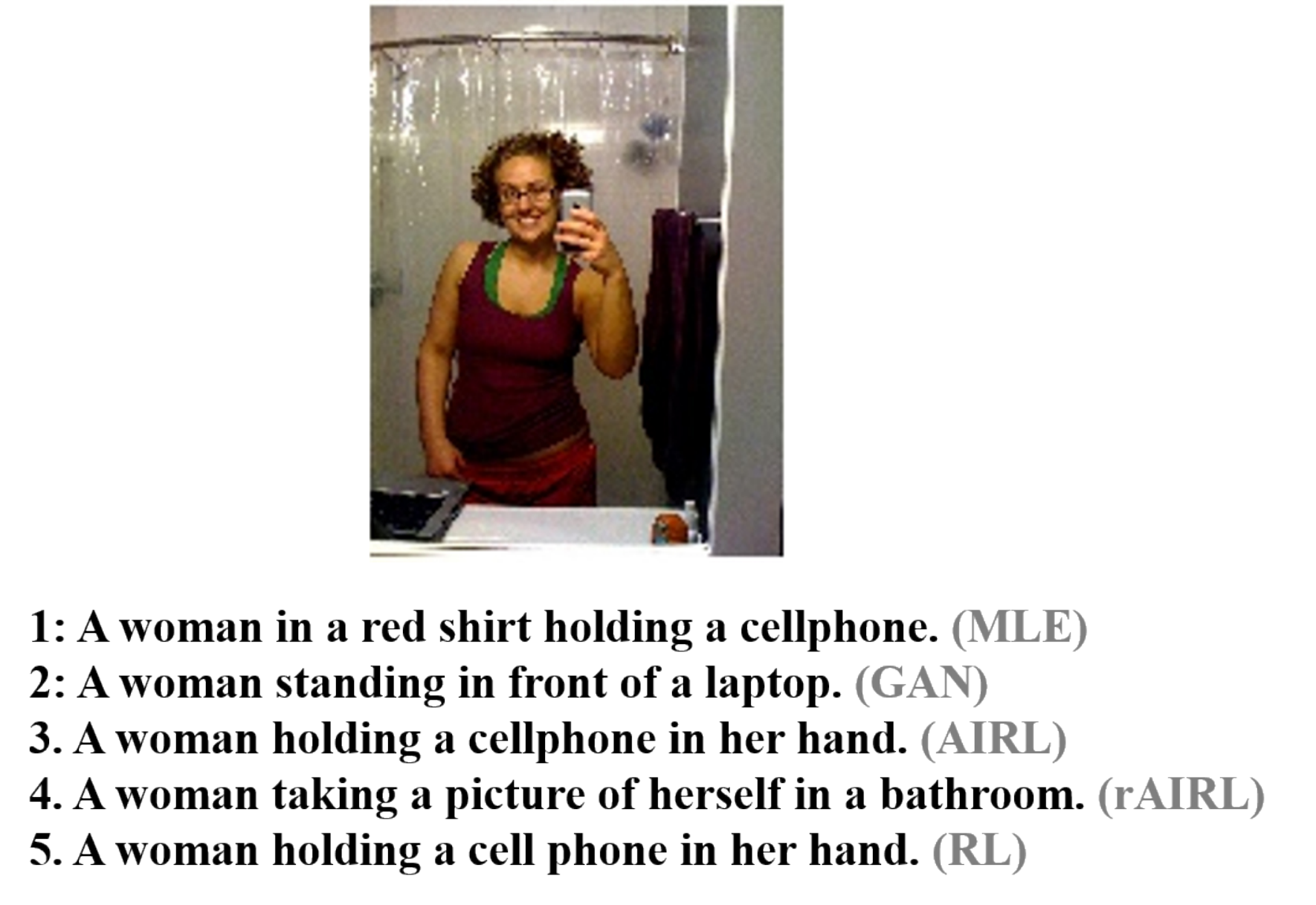}} 
  \caption{\small{An example of the images shown to the human evaluator in the human studies (methods marked in gray are not shown). The captions were produced by MLE, GAN, RL, AIRL and rAIRL methods in a randomized order. }}
  \label{fig:human}
\end{figure*}
\subsection{Evaluation on the Generated Captions}
In this section, we evaluated the generated captions mainly on three aspects: content correctness, diversity and grammar. Firstly, the results of the caption generator are compared with existing methods on the handcrafted evaluation metrics: BLEU4, CIDEr and SIPCE. However, since BLEU and CIDEr do not correlate well with human \citep{anderson16spice}, we choose other metrics to evaluate the captions in the following experiments. For a comprehensive evaluation, diversity and grammar are also considered as representation of the caption quality. Finally, results of the ablation studies are reported to show importance of each component of our algorithm in caption generation.
\paragraph{\textbf{Comparison with existing methods.}} Categorized by the loss functions, existing models are divided into three categories in Table \ref{tab:all}, and we chose recent proposed methods for comparison: Att2in \citep{Rennie2016Self}, G-GAN \citep{DaiL17Con}, NBT \citep{Lu2018Neural}, Up-Down \citep{peter2017bot} and GAN$_2$, GAN$_3$ \citep{DogninImproved}. Although some metrics based on $n$-gram overlapping (BLEU4, CIDEr) do not correlate well with human, their results are also reported in Table \ref{tab:all} for fair comparison. Among the adversarial methods (GAN category), our rAIRL performs the best on all metrics.\\
\indent
To test the generalizing ability of our algorithm, we also built our algorithm on the non-adversarial based models. The composite models are denoted with rAIRL +MLE and rAIRL+RL. In rAIRL+MLE, the conditional term is replaced by the cross-entropy loss of MLE; in rAIRL+RL, the RL loss is added into the loss function of the generator. In Table \ref{tab:all}, our rAIRL+MLE further improves the MLE baseline (i.e., Up-Down using MLE loss) on SPICE, whereas rAIRL+RL does not improve the RL baseline (i.e., Up-Down using RL loss) on these metrics. This is caused by the difficulty of normalizing the learned reward and the handcrafted reward to the same order of magnitude \citep{shelton2001balancing}. Although RL shows better performance on MS COCO by directly optimizing the handcrafted metric (CIDEr), we show in the following experiments that the overall quality of its generated descriptions is not as satisfying as it seems, especially on human evaluations and grammar. \\
\paragraph{\textbf{Content correctness.}} For a comprehensive evaluation of the content correctness, the results of both the handcrafted metrics and human studies are shown in Table \ref{tab:cd}. For the handcrafted metrics, we report scores of SPICE and the recently proposed CHAIR$_s$ and CHAIR$_i$ since they correlate well with human \citep{anderson16spice,rohrbach2018object}. SPICE computes similarity with the ground truth captions based on scene graph whilst CHIAIR$_s$ and CHIAR$_i$ indicate ratio of hallucinated objects. Compared with non-adversarial methods (i.e., MLE, RL), existing adversarial net (GAN) does not perform well on SPICE due to the reward ambiguity problem, whereas our rAIRL improves GAN (from $16.8$ to $18.7$) by disentangling reward for each word, and even outperforms RL (from $18.1$ to $18.7$) on the robust split. The lowest scores on CHIAIR$_s$ and CHIAR$_i$ suggest that object hallucination is less likely in rAIRL.\\
\indent
 As for the human evaluation, we randomly selected $500$ test images from the standard split and robust split of MS COCO, respectively. The worker was asked ``which caption is the best'' by given an image with five sentences generated from the adversarial and non-adversarial methods, as shown in Figure \ref{fig:human}. The worker was allowed but not encouraged to make multiple choices if he/she thinks these captions are equally correct. The order of captions produced by different methods was randomized. Following \citep{speak}, each image in the test set was evaluated by $5$ workers. \textbf{Human} in Table \ref{tab:cd} indicates the percentage of captions that are considered the best among the five methods. The descriptions generated by our rAIRL are considered the best for over $40\%$ images, whilst RL has the lowest scores that are less than $10\%$. The results of RL on human studies are almost contrary to its performance on the handcrafted metrics in Table \ref{tab:all}. This suggests that RL may optimize these metrics in an unintended way such that the scores are improved but the quality of caption is not. On the other hand, by self-learning a reward function, the proposed rAIRL has consistent performance on the human studies and the handcrafted metrics.
\begin{table}[h]
		\centering
		\caption{\small{Evaluation of the diversity on generated captions. The best score is in bold font and the second best score is underlined. \\}}
		\scalebox{0.8}[0.8]{
		\begin{tabular}{ccccc}\toprule
		 \multirow{3}{*}{Method } &\multicolumn{2}{c}{Standard Split} &\multicolumn{2}{c}{Robust Split} \cr \cmidrule(r){2-3} \cmidrule{4-5}
		 &Vocabulary  &Novel   &Vocabulary  &Novel\cr 
		 &Coverage &Sentence &Coverage &Sentence\cr \midrule
       MLE(Up-Down)   &\underline{12.4}   &49.7 &12.5    &58.8   \cr       
       RL(Up-Down)     &11.4  &\textbf{88.5}     &12.7   &\textbf{87.3} \cr
       GAN(Up-Down)    &13.4   &75.0 &15.3  &75.6 \cr
       AIRL(Up-Down)   &12.3   &67.3   &\underline{15.6}  &73.8  \cr
       rAIRL(Up-Down)  &\textbf{13.6}  &\underline{76.1}   &\textbf{15.8} &\underline{76.5}   \cr
		\bottomrule
		\end{tabular}		
		}	
		\label{tab:cd2}	
\end{table}
 \begin{figure*}[!t]
\centering
   \subfigure{ 
    \includegraphics[width=0.93\textwidth]{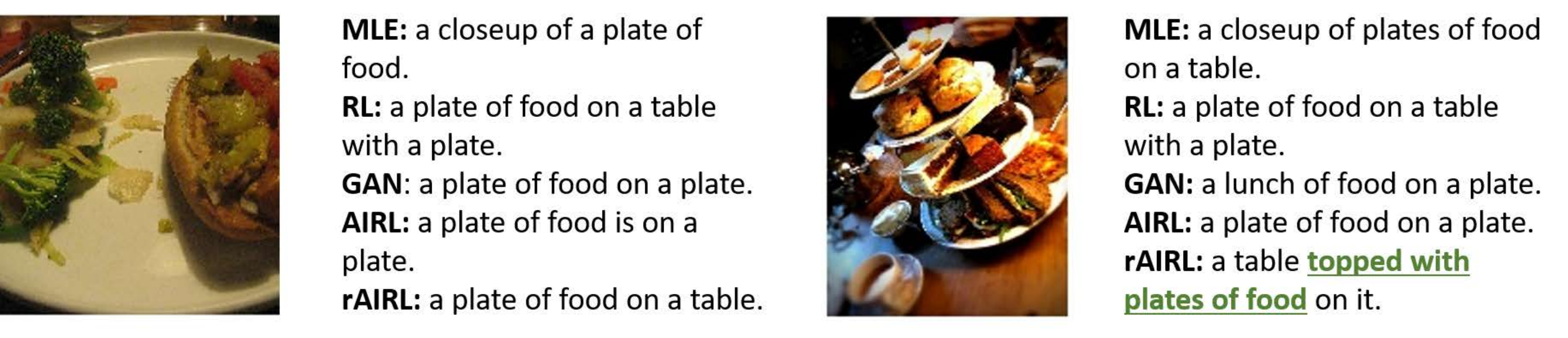}} 
    \rule{\textwidth}{0.5pt} 
    \subfigure{ 
    \includegraphics[width=0.93\textwidth]{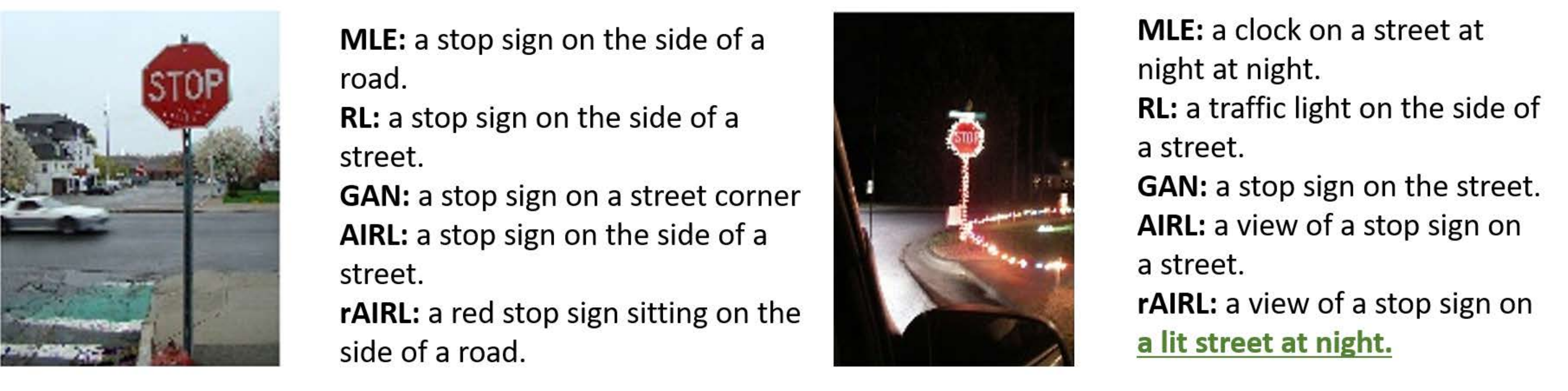}} 
  \caption{\small{Examples showing diversity of the captions. The left and right columns show pictures with similar content but different details. The proposed rAIRL successfully recognizes these differences and gives diverse captions.}}
    \label{fig:divisity}
\end{figure*}
\begin{table*}[!t]
		\centering
		\caption{\small{Percentage of different grammar errors found in the generated captions. Re represents Redundancy, AE is Agreement Error, AM denotes Article Misuse and IS is Incomplete Sentence.\\}}
		\begin{tabular}{ccccccccccc}\toprule
        \multirow{2}{*}{Method } &\multicolumn{5}{c}{Standard Split} &\multicolumn{5}{c}{Robust Split} \cr \cmidrule(r){2-6} \cmidrule{7-11}
        &Total &Re &AE &AM &IS &Total &Re &AE &AM &IS\cr \midrule
       MLE(Up-Down) &0.78  &0.04 &0.56  &0.14  &0.04  &\textbf{0.57} &0.04 &0.26 &0.16 &0.10  \cr
       RL(Up-Down)   &5.64 &0.90  &0  &3.36  &1.38  &4.67   &0.19 &0.02 &3.8 &0.69 \cr
       GAN(Up-Down) &1.24 &0.62 &0.18  &0.06 &0.38  &2.40 &1.10 &0.40  &0.26 &0.63\cr  
       AIRL(Up-Down)  &1.68 &0.04  &0.62 &0.70  &0.32  &1.20 &0.10 &0.27 &0.72 &0.12 \cr 
       rAIRL(Up-Down)   &\textbf{0.75} &0.14  &0.20 &0.21 &0.20  &\textbf{0.57} &0.14 &0.17  &0.16  &0.10 \cr
		\bottomrule
		\end{tabular}					
		\label{tab:gra}		
\end{table*}
\begin{table}[!t]
		\centering
		\small
		\caption{\small{Results of using different model architectures in our method.}}
		\scalebox{0.7}[0.7]{
        \begin{tabular}{ccccccc}\toprule     
	 \multirow{2}{*}{Method } &\multicolumn{3}{c}{Standard Split} &\multicolumn{3}{c}{Robust Split} \cr \cmidrule(r){2-4} \cmidrule{5-7}
      &BLEU4 &CIDEr &SPICE &BLEU4 &CIDEr &SPICE \cr \midrule
      Att2in &31.0 &101.3 &- &\textbf{31.5} &90.6 &17.7 \cr
      rAIRL(Att2in) &\textbf{31.3} &\textbf{105.2} &\textbf{19.9} &30.7 &\textbf{92.5} &\textbf{18.0}\cr \midrule   
       Up-Down &\textbf{36.2} &\textbf{113.5} &20.3 &\textbf{31.6} &92.0 &18.1\cr
       rAIRL(Up-Down) &33.8 &110.2 &\textbf{20.4} &30.2 &\textbf{93.7} &\textbf{18.7}\cr
		\bottomrule	
		\end{tabular}	
		}
		\label{tab:2}
\end{table}	
\paragraph{\textbf{Diversity.}} The diversity of captions is evaluated on a corpus level, indicating to what extent the generated captions of different images have diverse expressions. The results are presented in Table \ref{tab:cd2}. Vocabulary Coverage is the number of vocabularies of the generated captions over number of vocabularies of the ground truth captions. Novel Sentence indicates the ratio of sentences that do not appear in the training set. The fact that RL has high ratio of novel sentence ($88.5\%$/$87.3\%$) but low vocabulary coverage ($11.4\%$/$12.7\%$) suggests that it uses high-frequency words (such as ``in a'', ``of a'') to reconstruct captions that appear to be different from the training set \citep{me}. Our rAIRL improves AIRL on the diversity metrics and outperforms other learning methods on vocabulary coverage, indicating its capability of generating diverse descriptions on a corpus level. Figure \ref{fig:divisity} gives two examples showing diversity of the generated caption. The proposed rAIRL recognizes notable differences between two similar images and give diverse descriptions for each image.
\begin{table*}[!t]
		\centering
		\caption{\small{Ablation methods of rAIRL. ``term1'' is the constant term in Eq. (\ref{eq:lt}) and ``term2'' is the conditional term in Eq. (\ref{eq:gen}). GE denotes grammar error rate.VC denotes vocabulary coverage and NS is the ratio of novel sentence.\\}}
		\scalebox{0.9}[0.9]{
		\begin{tabular}{ccccccccccccc}\toprule
	 \multirow{2}{*}{Method } &\multicolumn{6}{c}{Standard Split} &\multicolumn{6}{c}{Robust Split} \cr \cmidrule(r){2-7} \cmidrule{8-13}
      &SPICE  &CHAIR$_s$ &CHAIR$_i$ &VC &NS &GE  &SPICE  &CHAIR$_s$ &CHAIR$_i$ &VC &NS &GE\cr \midrule
       rAIRL(Up-Down, w/o term1) &18.8  &10.5 &8.2 &12.8 &73.5 &1.07 &17.0 &19.9 &17.5  &14.1 &71.6 &0.95 \cr
       rAIRL(Up-Down, w/o term2) &19.3  &9.4 &7.4 &12.2 &71.3 &0.83  &17.9  &18.9 &15.8  &13.7 &62.4 &0.72\cr
       rAIRL(Up-Down)  &\textbf{20.4}   &\textbf{7.2} &\textbf{5.5}  &\textbf{13.6}  &\textbf{76.1} &\textbf{0.75} &\textbf{18.7} &\textbf{17.1} &\textbf{14.3}  &\textbf{15.8} &\textbf{76.5} &\textbf{0.57} \cr
		\bottomrule
		\end{tabular}		
		}
		\label{tab:abl}		
\end{table*}
\begin{figure*}[!t]
\centering
   \subfigure{ 
    \includegraphics[width=0.85\textwidth]{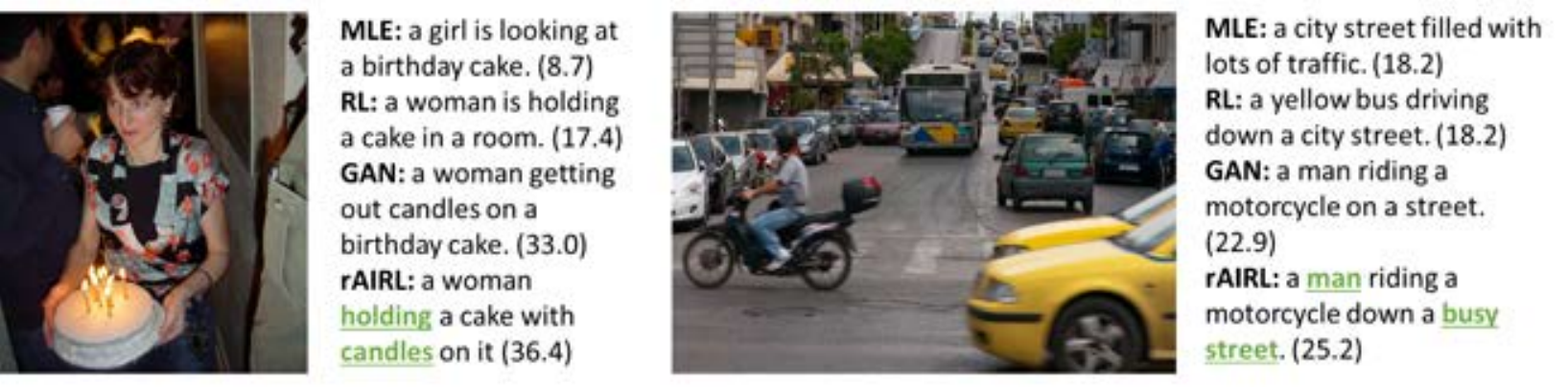}} 
    
    \subfigure{ 
    \includegraphics[width=0.85\textwidth]{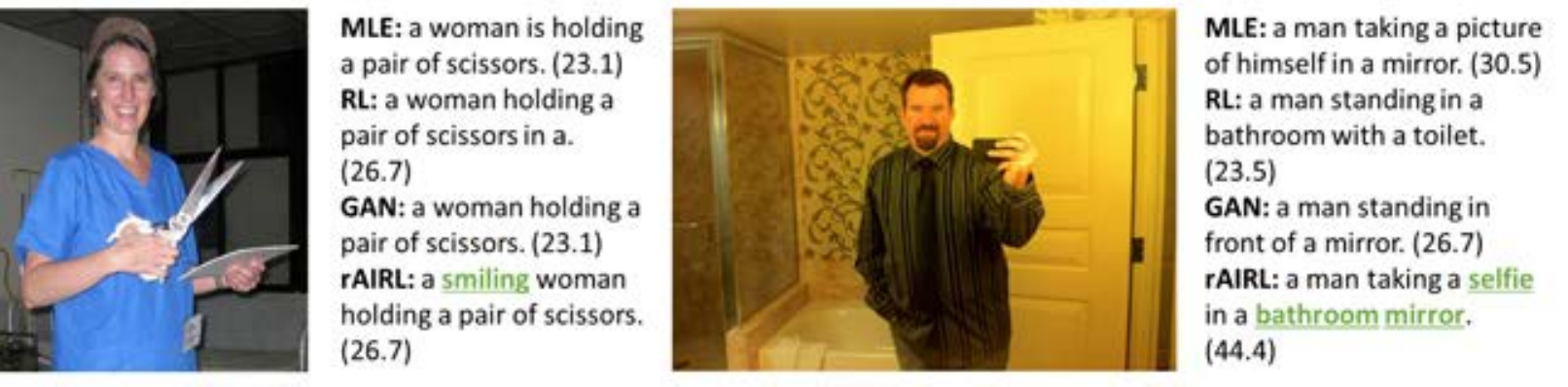}} 
    
    \subfigure{ 
    \includegraphics[width=0.85\textwidth]{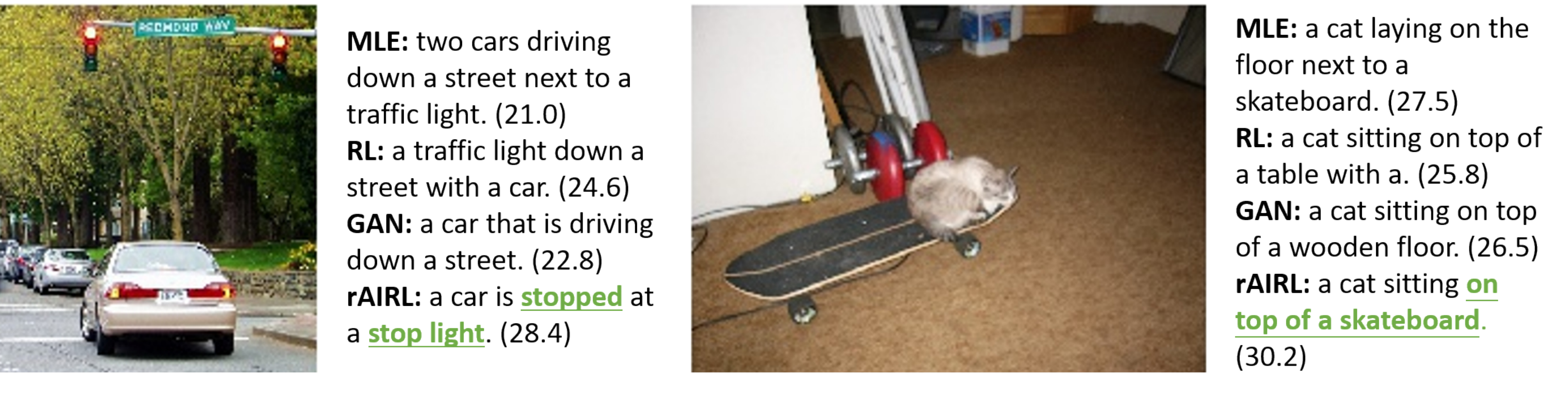}} 
    
  \caption{\small{Captions produced by different methods from the test set (standard split). Beside each caption we report SPICE score. Captions generated by rAIRL are correct and human-like in these examples.}}
  \label{fig:vis1}
\end{figure*}
\begin{figure*}[!t]
\centering
    \includegraphics[width=0.85\textwidth]{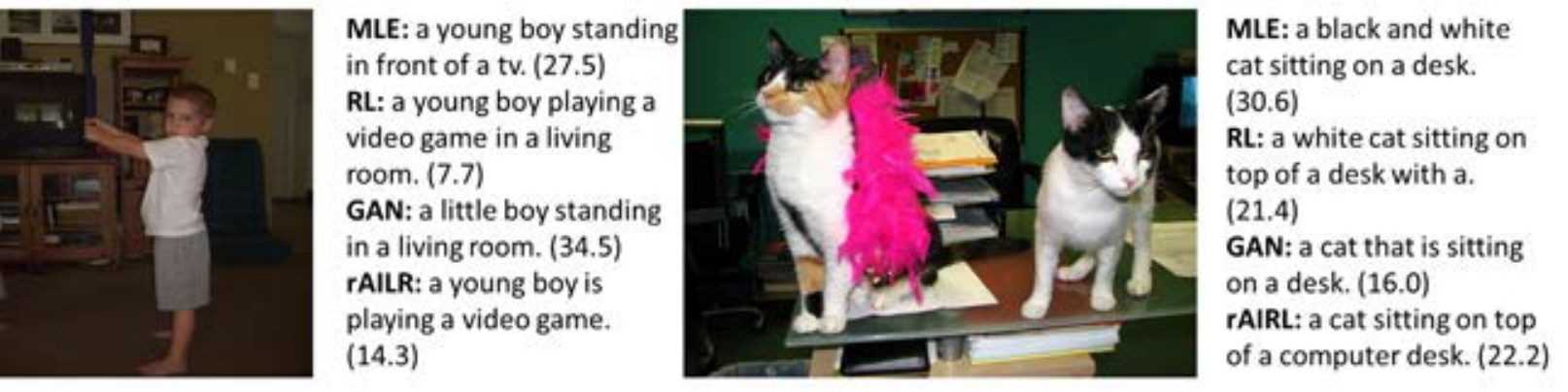}
  \caption{\small{Failed examples of  rAIRL. The objects and relations are not correctly recognized in these pictures. }}
    \label{fig:vis2}
\end{figure*}
\paragraph{\textbf{Grammar.}} We used LanguageTool \footnote{\url{https://languagetool.org}} to check grammar of the generated captions. Table \ref{tab:gra} shows percentage of sentences that have grammar errors found by LanguageTool: 1) \emph{Redundancy} means repeated phrases in a sentence; 2) \emph{Agreement Error} means subject-verb agreement error, such as ``people is''; 3) \emph{Article Misuse} denotes inappropriate usage of indefinite articles, such as using ``a'' before uncountable nouns or plural words;  4) \emph{Incomplete Sentence} refers to incomplete sentence that lacks a subject. We found captions produced by RL have the most grammar errors ($5.64\%$ on the standard split and $4.67\%$ on the robust split), especially the Article Misuse. On the other hand, by approximating the true data distribution of each word in the sentence, rAIRL and MLE have the least grammar errors among all learning methods ($0.75\%/0.78\%$ on the standard split and $0.57\%/0.57\%$ on the robust split)). We also noticed that each method except rAIRL is biased towards a particular type of grammar error: agreement error in MLE, article misuse in RL, redundancy in GAN, article misuse in AIRL. On both splits, our rAIRL does not appear to be biased towards a specific type of these grammar errors.
\paragraph{\textbf{Ablation studies.}} Theoretically, our algorithm is model-agnostic since it is independent of the design of model architecture. Therefore, we compare the results of using Att2in \citep{Rennie2016Self} and Up-Down \citep{peter2017bot} model architectures in Table \ref{tab:2}, respectively. We report the metrics used in the original paper for fair comparison. The proposed rAIRL mainly improves SPICE on both architectures.\\
\indent
We conducted another ablation experiment to understand the importance of each component of our algorithm in caption generation. Specifically, the \emph{constant term} in Eq. (\ref{eq:lt}) and the \emph{conditional term} in Eq. (\ref{eq:gen}) is removed, respectively. Scores of all the evaluation techniques mentioned above are presented in Table \ref{tab:abl}. All the scores have a drop after removing either one of the terms. Comparing these two terms, the \emph{constant term} seems more important in recognizing objects and relations in the image since removing it has larger drop on SPICE. The lager drop on vocabulary coverage and ratio of novel sentence in the second row indicates that the \emph{conditional term} plays a significant role in increasing the diversity of the generated captions.
\paragraph{\textbf{Visualized examples.}} A few examples of the generated captions produced by different methods are shown in Figures \ref{fig:vis1} and \ref{fig:vis2}. We compare the captioning results of rAIRL with three other methods: MLE, RL and GAN. Figure \ref{fig:vis1} gives successful examples, especially on captioning relations between objects. Figure \ref{fig:vis2} shows failed examples, where objects and relations are not correctly recognized by the captioning model.
\paragraph{\textbf{Summary.}} Through extensive experiments on caption generation, we proved that the proposed rAIRL constantly performs well on both splits of MS COCO. Compared with RL, rAIRL optimizes the learned reward instead of the handcrafted metrics, and is capable of producing qualitative captions with few grammar errors. As an adversarial algorithm, rAIRL enhances GAN by disentangling compact reward for each word in the caption and improves AIRL by shifting the generator towards Nash equilibrium. 

\section{Conclusion}
In this paper, we address the reward ambiguity problem in image captioning and propose a refined Adversarial Inverse Reinforcement Learning (rAIRL) method that solves the problem by disentangling reward for each word in a sentence. Moreover, it achieves stable adversarial training by refining the loss function to shift the generator towards Nash equilibrium, and mode control technique is incorporated to mitigate mode collapse. It is demonstrated that our method can learn compact reward through extensive experiments on MS COCO and Flickr30K.



\normalsize
\bibliography{myBib}

\begin{thebibliography}{38}
\providecommand{\natexlab}[1]{#1}
\providecommand{\url}[1]{\texttt{#1}}
\expandafter\ifx\csname urlstyle\endcsname\relax
  \providecommand{\doi}[1]{doi: #1}\else
  \providecommand{\doi}{doi: \begingroup \urlstyle{rm}\Url}\fi

\bibitem[Vinyals et~al.(2015)Vinyals, Toshev, Bengio, and
  Erhan]{Vinyals2015Show}
Oriol Vinyals, Alexander Toshev, Samy Bengio, and Dumitru Erhan.
\newblock Show and tell: A neural image caption generator.
\newblock In \emph{\emph{{IEEE} conference on computer vision and pattern
  recognition}}, pages 3156--3164, 2015.

\bibitem[Xu et~al.(2015)Xu, Ba, Kiros, Cho, Courville, Salakhudinov, Zemel, and
  Bengio]{xu2015show}
Kelvin Xu, Jimmy Ba, Ryan Kiros, Kyunghyun Cho, Aaron Courville, Ruslan
  Salakhudinov, Rich Zemel, and Yoshua Bengio.
\newblock Show, attend and tell: Neural image caption generation with visual
  attention.
\newblock In \emph{\emph{International conference on machine learning}}, pages
  2048--2057, 2015.

\bibitem[Dai et~al.(2017)Dai, Lin, Urtasun, and Fidler]{DaiLUF17}
Bo~Dai, Dahua Lin, Raquel Urtasun, and Sanja Fidler.
\newblock Towards diverse and natural image descriptions via a conditional
  {GAN}.
\newblock In \emph{\emph{International conference on computer vision}}, pages
  2989--2998, 2017.

\bibitem[Rennie et~al.(2017)Rennie, Marcheret, Mroueh, Ross, and
  Goel]{Rennie2016Self}
Steven~J. Rennie, Etienne Marcheret, Youssef Mroueh, Jerret Ross, and Vaibhava
  Goel.
\newblock Self-critical sequence training for image captioning.
\newblock In \emph{\emph{{IEEE} conference on computer vision and pattern
  recognition}}, pages 1179--1195, 2017.

\bibitem[Liu et~al.(2017)Liu, Zhu, Ye, Guadarrama, and Murphy]{si2017improve}
Siqi Liu, Zhenhai Zhu, Ning Ye, Sergio Guadarrama, and Kevin Murphy.
\newblock Improved image captioning via policy gradient optimization of
  {SPIDE}r.
\newblock In \emph{\emph{International conference on computer vision}}, pages
  873--881, 2017.

\bibitem[Chen et~al.(2019)Chen, Mu, Xiao, Ye, Wu, and Ju]{chen2019improving}
Chen Chen, Shuai Mu, Wanpeng Xiao, Zexiong Ye, Liesi Wu, and Qi~Ju.
\newblock Improving image captioning with conditional generative adversarial
  nets.
\newblock In \emph{\emph{AAAI conference on artificial intelligence}}, pages
  8142--8150, 2019.

\bibitem[Li et~al.(2019{\natexlab{a}})Li, Chen, and Liu]{me}
Nannan Li, Zhenzhong Chen, and Shan Liu.
\newblock Meta learning for image captioning.
\newblock In \emph{\emph{AAAI conference on artificial intelligence}}, pages
  8626--8633, 2019{\natexlab{a}}.

\bibitem[Shetty et~al.(2017)Shetty, Rohrbach, Hendricks, Fritz, and
  Schiele]{speak}
Rakshith Shetty, Marcus Rohrbach, Lisa~Anne Hendricks, Mario Fritz, and Bernt
  Schiele.
\newblock Speaking the same language: Matching machine to human captions by
  adversarial training.
\newblock In \emph{\emph{International conference on computer vision}}, pages
  4155--4164, 2017.

\bibitem[Dognin et~al.(2019)Dognin, Melnyk, Mroueh, Ross, and
  Sercu]{DogninImproved}
Pierre~L. Dognin, Igor Melnyk, Youssef Mroueh, Jerret Ross, and Tom Sercu.
\newblock Adversarial semantic alignment for improved image captions.
\newblock In \emph{\emph{{IEEE} conference on computer vision and pattern
  recognition}}, pages 10463--10471, 2019.

\bibitem[Goodfellow et~al.(2014)Goodfellow, Pouget-Abadie, Mirza, Xu,
  Warde-Farley, Ozair, Courville, and Bengio]{goodfellow2014generative}
Ian Goodfellow, Jean Pouget-Abadie, Mehdi Mirza, Bing Xu, David Warde-Farley,
  Sherjil Ozair, Aaron Courville, and Yoshua Bengio.
\newblock Generative adversarial nets.
\newblock In \emph{\emph{Advances in neural information processing systems }},
  pages 2672--2680, 2014.

\bibitem[Ng et~al.(1999)Ng, Harada, and Russell]{ng1999}
Andrew Ng, Daishi Harada, and Stuart Russell.
\newblock Policy invariance under reward transformations: Theory and
  application to reward shaping.
\newblock In \emph{\emph{ International Conference on Machine Learning}}, pages
  278--287, 1999.

\bibitem[Fu et~al.(2018)Fu, Luo, and Levine]{Fu2018LearningRobust}
Justin Fu, Katie Luo, and Sergey Levine.
\newblock Learning robust rewards with adversarial inverse reinforcement
  learning.
\newblock In \emph{\emph{International conference on learning
  representations}}, 2018.

\bibitem[Mescheder and Geiger(2017)]{the-numerics-of-gans}
Lars~M. Mescheder and Andreas Geiger.
\newblock The numerics of gans.
\newblock In \emph{\emph{Advances in neural information processing systems }},
  pages 1825--1835, 2017.

\bibitem[Mirza and Osindero(2014)]{MirzaO14}
Mehdi Mirza and Simon Osindero.
\newblock Conditional generative adversarial nets.
\newblock \emph{\emph{arXiv preprint arXiv:1411.1784}}, 2014.

\bibitem[Cui et~al.(2018)Cui, Yang, Veit, Huang, and Belongie]{cui2018learning}
Yin Cui, Guandao Yang, Andreas Veit, Xun Huang, and Serge~J Belongie.
\newblock Learning to evaluate image captioning.
\newblock In \emph{\emph{{IEEE} conference on computer vision and pattern
  recognition}}, pages 5804--5812, 2018.

\bibitem[Sharif et~al.(2018)Sharif, White, Bennamoun, and
  Shah]{sharif2018learning-based}
Naeha Sharif, Lyndon White, Mohammed Bennamoun, and Syed Afaq~Ali Shah.
\newblock Learning-based composite metrics for improved caption evaluation.
\newblock In \emph{\emph{ACL}}, pages 14--20, 2018.

\bibitem[Lu et~al.(2017)Lu, Xiong, Parikh, and Socher]{Lu2017Knowing}
Jiasen Lu, Caiming Xiong, Devi Parikh, and Richard Socher.
\newblock Knowing when to look: Adaptive attention via a visual sentinel for
  image captioning.
\newblock In \emph{\emph{ {IEEE} conference on computer vision and pattern
  recognition}}, pages 3242--3250, 2017.

\bibitem[Yao et~al.(2018)Yao, Pan, Li, and Mei]{explore2018yao}
Ting Yao, Yingwei Pan, Yehao Li, and Tao Mei.
\newblock Exploring visual relationship for image captioning.
\newblock In \emph{\emph{European conference on computer vision}}, pages
  711--727, 2018.

\bibitem[Ren et~al.(2017)Ren, Wang, Zhang, Lv, and Li]{Ren_2017_CVPR}
Zhou Ren, Xiaoyu Wang, Ning Zhang, Xutao Lv, and Li-Jia Li.
\newblock Deep reinforcement learning-based image captioning with embedding
  reward.
\newblock In \emph{\emph{{IEEE} conference on computer vision and pattern
  recognition}}, pages 1151--1159, 2017.

\bibitem[Lu et~al.(2018)Lu, Yang, Batra, and Parikh]{Lu2018Neural}
Jiasen Lu, Jianwei Yang, Dhruv Batra, and Devi Parikh.
\newblock Neural baby talk.
\newblock In \emph{\emph{ {IEEE} conference on computer vision and pattern
  recognition}}, pages 7219--7228, 2018.

\bibitem[Wang et~al.(2018)Wang, Chen, Wang, and Wang]{WangNo18}
Xin Wang, Wenhu Chen, Yuan{-}Fang Wang, and William~Yang Wang.
\newblock No metrics are perfect: Adversarial reward learning for visual
  storytelling.
\newblock In \emph{\emph{ACL}}, pages 899--909, 2018.

\bibitem[Li et~al.(2019{\natexlab{b}})Li, Kiseleva, and de~Rijke]{ziming19diag}
Ziming Li, Julia Kiseleva, and Maarten de~Rijke.
\newblock Dialogue generation: From imitation learning to inverse reinforcement
  learning.
\newblock In \emph{\emph{AAAI conference on artificial intelligence}}, pages
  6722--6729, 2019{\natexlab{b}}.

\bibitem[Shi et~al.(2018)Shi, Chen, Qiu, and Huang]{zhan18towards}
Zhan Shi, Xinchi Chen, Xipeng Qiu, and Xuanjing Huang.
\newblock Towards diverse text generation with inverse reinforcement learning.
\newblock In \emph{\emph{International joint conference on artficial
  intelligence}}, pages 4361--4367, 2018.

\bibitem[Finn et~al.(2016)Finn, Christiano, Abbeel, and
  Levine]{Finn2016AConnection}
Chelsea Finn, Paul~F. Christiano, Pieter Abbeel, and Sergey Levine.
\newblock A connection between generative adversarial networks, inverse
  reinforcement learning, and energy-based models.
\newblock In \emph{\emph{arXiv preprint arXiv:1611.03852}}, 2016.

\bibitem[Sutton and Barto(1998)]{sutton1998reinforcement}
Richard~S Sutton and Andrew~G Barto.
\newblock \emph{\emph{Reinforcement learning: An introduction}}.
\newblock MIT press Cambridge, 1998.

\bibitem[Peng et~al.(2019)Peng, Kanazawa, Toyer, Abbeel, and
  Levine]{peng2018variational}
Xue~Bin Peng, Angjoo Kanazawa, Sam Toyer, Pieter Abbeel, and Sergey Levine.
\newblock Variational discriminator bottleneck: {I}mproving imitation learning,
  inverse {RL}, and {GAN}s by constraining information flow.
\newblock In \emph{\emph{International conference on learning
  representations}}, 2019.

\bibitem[Chen et~al.(2015)Chen, Fang, Lin, Vedantam, Gupta, Dollár, and
  Zitnick]{capeval2015}
Xinlei Chen, Hao Fang, Tsung-Yi Lin, Ramakrishna Vedantam, Saurabh Gupta, Piotr
  Dollár, and C.~Lawrence Zitnick.
\newblock Microsoft \protect{COCO} captions: Data collection and evaluation
  server.
\newblock 2015.

\bibitem[Young et~al.(2014)Young, Lai, Hodosh, and Hockenmaier]{Young2014From}
P.~Young, A.~Lai, M.~Hodosh, and J.~Hockenmaier.
\newblock From image descriptions to visual denotations: New similarity metrics
  for semantic inference over event descriptions.
\newblock In \emph{\emph{ACL}}, 2014.

\bibitem[Karpathy and Fei-Fei(2015)]{Karpathy2015Deep}
Andrej Karpathy and Li~Fei-Fei.
\newblock Deep visual-semantic alignments for generating image descriptions.
\newblock In \emph{\emph{{IEEE} conference on computer vision and pattern
  recognition}}, pages 3128--3137, 2015.

\bibitem[Kingma and Ba(2014)]{kingma2014adam}
Diederik~P. Kingma and Jimmy Ba.
\newblock Adam: A method for stochastic optimization.
\newblock In \emph{\emph{International conference on learning
  representations}}, 2014.

\bibitem[Anderson et~al.(2018)Anderson, He, Buehler, Teney, Johnson, Gould, and
  Zhang]{peter2017bot}
Peter Anderson, Xiaodong He, Chris Buehler, Damien Teney, Mark Johnson, Stephen
  Gould, and Lei Zhang.
\newblock Bottom-up and top-down attention for image captioning and visual
  question answering.
\newblock In \emph{\emph{{IEEE} conference on computer vision and pattern
  recognition}}, pages 6077--6086, 2018.

\bibitem[Dai and Lin(2017)]{DaiL17Con}
Bo~Dai and Dahua Lin.
\newblock Contrastive learning for image captioning.
\newblock In \emph{\emph{Advances in neural Information processing systems}},
  pages 898--907, 2017.

\bibitem[Aditya et~al.(2017)Aditya, Yang, Baral, Aloimonos, and
  Fermüller]{Aditya2017}
Somak Aditya, Yezhou Yang, Chitta Baral, Yiannis Aloimonos, and Cornelia
  Fermüller.
\newblock Image understanding using vision and reasoning through scene
  description graph.
\newblock \emph{\emph{Computer Vision and Image Understanding }}, 173:\penalty0
  33--45, 2017.

\bibitem[Anderson et~al.(2016)Anderson, Fernando, Johnson, and
  Gould]{anderson16spice}
Peter Anderson, Basura Fernando, Mark Johnson, and Stephen Gould.
\newblock {SPICE:} semantic propositional image caption evaluation.
\newblock In \emph{\emph{European conference on computer vision}}, pages
  382--398, 2016.

\bibitem[Pennington et~al.(2014)Pennington, Socher, and
  Manning]{pennington2014glove}
Jeffrey Pennington, Richard Socher, and Christopher Manning.
\newblock Glove: Global vectors for word representation.
\newblock In \emph{\emph{Empirical methods in natural language processing}},
  pages 1532--1543, 2014.

\bibitem[Johnson et~al.(2015)Johnson, Krishna, Stark, Li, Shamma, Bernstein,
  and Feifei]{johnson2015image}
Justin Johnson, Ranjay Krishna, Michael Stark, Lijia Li, David~A Shamma,
  Michael~S Bernstein, and Li~Feifei.
\newblock Image retrieval using scene graphs.
\newblock In \emph{\emph{{IEEE} conference on computer vision and pattern
  recognition}}, pages 3668--3678, 2015.

\bibitem[Rohrbach et~al.(2018)Rohrbach, Hendricks, Burns, Darrell, and
  Saenko]{rohrbach2018object}
Anna Rohrbach, Lisa~Anne Hendricks, Kaylee Burns, Trevor Darrell, and Kate
  Saenko.
\newblock Object hallucination in image captioning.
\newblock In \emph{\emph{Empirical methods in natural language processing}},
  pages 4035--4045, 2018.

\bibitem[Shelton~Christian(2001)]{shelton2001balancing}
R~Shelton~Christian.
\newblock Balancing multiple sources of reward in reinforcement learning.
\newblock In \emph{\emph{Advances in neural information processing systems }},
  pages 1082--1088, 2001.

\end{thebibliography}


\end{document}